\documentclass[5p,times,authoryear]{RIAIarticle}

\usepackage{crc_RIAI}

\usepackage[spanish]{babel}     
\usepackage[utf8]{inputenc} 
\addto\captionsspanish{%
}

\usepackage{amsmath}            
\usepackage{epstopdf}           
\usepackage{flushend}           

\usepackage{graphics} 
\usepackage{epsfig} 
\usepackage{bmpsize}  
\usepackage{glossaries}
\newacronym{rnn}{RNN}{Recurrent Neural Network}
\newacronym{resnet}{ResNet}{Residual Neural Network}
\newacronym{mlp}{MLP}{Multilayer Perceptron}
\newacronym{svm}{SVM}{Support Vector Machine}
\newacronym{crnn}{CRNN}{Convolutional Recurrent Neural Network}
\newacronym[plural=NNs, firstplural=Neural Networks (NNs)]{nn}{NN}{Neural Network}
\newacronym[plural=CNNs, firstplural=Convolutional Neural Networks 
(CNNs)]{cnn}{CNN}{Convolutional Neural Network}
\newacronym[plural=RoIs, firstplural=Regions of Interest (RoIs)]{roi}{RoI}{Region of Interest}
\newacronym{ros}{ROS}{Robot Operative System}
\newacronym{bb}{BB}{Bounding Box}
\newacronym{fpn}{FPN}{Feature Pyramid Network}
\newacronym{fps}{FPS}{Frames Per Second}
\newacronym{dcnn}{DCNN}{Deep Convolutional Neural Network}
\newacronym{gpc}{GPC}{Gaussian Process Classification}
\newacronym{socn}{SOCN}{Stacked Object Classification Network}
\newacronym{pcb}{PCB}{Printed Circuit Board}

\volume{00}

\firstpage{1}

\journalname{Revista Iberoamericana de Automática e Informática Industrial}

\runauth{Casta\~{n}o-Amor\'os, J.}

\jid{RIAI}

\jnltitlelogo{}

\usepackage{amssymb}

\usepackage[figuresright]{rotating}

\usepackage{subfigure}
\usepackage{float} 
\usepackage{multicol} 
\usepackage[most]{tcolorbox}

\begin{document}

\begin{frontmatter}


\title{Manipulación visual-táctil para la recogida de residuos dom\'esticos en exteriores}


\author[First,Third]{Julio Casta\~{n}o-Amor\'os}

\author[First,Third]{Ignacio de Loyola P\'aez-Ubieta}

\author[First,Second]{Pablo Gil\corref{cor1}}

\author[First,Second]{Santiago Timoteo Puente}

\cortext[cor1]{Autor para correspondencia: pablo.gil@ua.es
\\
Attribution-NonCommercial-ShareAlike 4.0 International (CC BY-NC-SA 4.0)
}

\address[First]{Instituto Universitario de Investigación Informática, Universidad de Alicante, Campus de San Vicente del Raspeig s/n, 03690, Alicante, España.}
\address[Second]{Departamento de Física, Ingeniería de Sistemas y Teoría de la Señal, Universidad de Alicante, Campus de San Vicente del Raspeig s/n, 03690, Alicante, España.}
\address[Third]{Universidad Miguel Hernández, 03202, Elche, España.}

\begin{comocitar}
\begin{center}
\begin{tcolorbox}[frame hidden, width=\linewidth*2, colframe=white, colback=black!8!white]
\tocitearticle{Casta\~{n}o-Amor\'os, J., P\'aez-Ubieta, I. de L., Gil, P., Puente, S. 2022. Visual-tactile manipulation to collect household waste in outdoor. Revista Iberoamericana de Autom\'atica e Inform\'atica Industrial 00, 1-5. https://doi.org/10.4995/riai.2020.7133}
\end{tcolorbox}
\end{center}
\end{comocitar}

\begin{abstract}
Este artículo presenta un sistema de percepción orientado a la manipulación robótica, capaz de asistir en tareas de navegación, clasificación y recogida de residuos domésticos en exterior. El sistema está compuesto de sensores táctiles ópticos, cámaras RGBD y un LiDAR. Éstos se integran en una plataforma móvil que transporta un robot manipulador con pinza. El sistema consta de tres módulos software, dos visuales y uno táctil. Los módulos visuales implementan arquitecturas \glspl{cnn} para la localización y reconocimiento de residuos sólidos, además de estimar puntos de agarre. El módulo táctil, también basado en \glspl{cnn} y procesamiento de imagen, regula la apertura de la pinza para controlar el agarre a partir de información de contacto. Nuestra propuesta tiene errores de localización entorno al 6 \%, una precisión de reconocimiento del 98 \%, y garantiza estabilidad de agarre el 91 \% de las veces. Los tres módulos trabajan en tiempos inferiores a los 750 ms.

\end{abstract}


\begin{englishtitle}
\noindent \textbf{Visual-tactile manipulation to collect household waste in outdoor}
\end{englishtitle}

\begin{abstractIng}
This work presents a perception system applied to robotic manipulation, that is able to assist in navegation, household waste classification and collection in outdoor environments. This system is made up of optical tactile sensors, RGBD cameras and a LiDAR. These sensors are integrated on a mobile platform with a robot manipulator and a robotic gripper. Our system is divided in three software modules, two of them are vision-based and the last one is tactile-based. The vision-based modules use \glspl{cnn} to localize and recognize solid household waste, together with the grasping points estimation. The tactile-based module, which also uses \glspl{cnn} and image processing, adjusts the gripper opening to control the grasping from touch data. Our proposal achieves localization errors around 6\%, a recognition accuracy of 98\% and ensures the grasping stability the 91\% of the attempts. The sum of runtimes of the three modules is less than 750 ms.
\end{abstractIng}

\begin{keywordIng}
Visual detection \sep Object recognition \sep Object location\sep Tactile perception \sep Robotic manipulation 
\end{keywordIng}

\end{frontmatter}

\begin{multicols}{2}   


\section{Introducci\'on}
En la actualidad, uno de los desafíos mundiales está relacionado con la digitalización en el cuidado de nuestro entorno, y más en concreto en su aplicación en áreas municipalizadas. Basta hacer notar que, en los últimos años, en España se generaron una media de 483,7 Kilos/hab en 2019, último año registrado según el Instituto Nacional de Estadística (INE), es decir 17.764 toneladas. Solamente entre residuos urbanos no peligrosos y recuperables de vidrio, cartón y plástico se generaron 2.3 millones en el territorio nacional, siendo una tendencia al alza el número de residuos procedentes de basura doméstica. Muchos de estos residuos acaban fuera de sus depósitos de recolección, en parques, recintos abiertos y calles de áreas urbanizadas. En este trabajo se expone una solución robotizada que trata de paliar la acumulación de residuos domésticos en exteriores, fuera de los lugares y depósitos de recolección destinados para ello. En concreto, se presenta un sistema sensorial enfocado a facilitar la recogida automatizada y organizada de desechos no orgánicos y no peligrosos, como botellas de plástico, latas, cajas, etc.

Este artículo se organiza de la siguiente manera. En la Sección 2, se exponen trabajos previos relacionados con el reconocimiento de residuos para su  manipulación en distintos escenarios. Posteriormente, en la Sección 3, se describe la plataforma robótica empleada para percibir el entorno, los objetos presentes en él y ejecutar así, las tareas de manipulación y recogida de residuos en exteriores. A continuación, en la Sección 4, se presentan los métodos y algoritmos implementados para detección, localización y cálculo de agarre a partir de información visual. Y en la Sección 5, se describen los métodos y algoritmos táctiles desarrollados para llevar a cabo un agarre estable y exitoso en tareas de levantar y soltar. Finalmente, en la última sección, se muestran y discuten resultados experimentales llevados a cabo en escenarios reales.


\section{Trabajos relacionados}

Los avances en procesadores gráficos y en técnicas computacionales que se han producido recientemente han repercutido directamente en mejorar el procesamiento e interpretación de imágenes de objetos, tanto en escenarios de interior como de exterior, con condiciones de iluminación cambiantes y con factores de incertidumbre en pose, forma, textura, etc. Los métodos de aprendizaje máquina, y especialmente aquellos basados en redes neuronales, son los grandes beneficiados, ya que han visto reducir drásticamente los tiempos de entrenamiento e inferencia, permitiendo manejar grandes volúmenes de datos y muchas casuísticas distintas de objetos y escenarios. Este boom computacional está permitiendo, por un lado, tratar de abordar el reconocimiento de residuos en la industria de reciclado como un problema de clasificación automática de imágenes. Así en \citep{bircan2018}, \citep{patrizi2021} y \citep{Vo2019}  se evalúa, para esa tarea, el uso de distintas versiones de conocidas \glspl{cnn} (ResNet, Inception, Xception, MobileNet, etc.) y, se proponen ligeras modificaciones en las últimas capas para clasificar objetos, que generalmente se presentan aislados, apoyados sobre superficies totalmente planas, homogéneas, y sin oclusiones. Por ejemplo, en \citep{bircan2018} se clasifican las imágenes en muestras de papel, vidrio y metal, mientras que en \citep{Vo2019} se clasifican representaciones orgánicas, inorgánicas o residuos médicos. En \citep{patrizi2021} se generan nuevas muestras de residuos a partir de las bases de datos empleadas en los anteriores trabajos. Para hacerlo, los autores aplican técnicas de aumentado de datos por sustitución del fondo de las imágenes de los objetos con el objetivo de mejorar la generalización en la clasificación. 

También, ha habido intentos de implementar CNNs más ligeras con pocas capas para reducir tiempos de entrenamiento e inferencia como en \citep{altikat2022} con bastante peor tasa de reconocimiento, o incluso de embeber este tipo de CNNs en sistemas de CPU integrada del tipo Raspberry Pi 4 como en \citep{bowen2021}, para así diseñar sistemas portables y poder controlar actuadores desde la salida del reconocimiento.

Por otro lado, en el estado del arte, también se ha tratado de enfocar el problema de reconocimiento de residuos como una tarea de detección de objetos, como muestran trabajos como \citep{feng2021} y \citep{kiyokawa2021}. A diferencia de abordarlo como un problema de clasificación, hacerlo como uno de detección ayuda a extraer características que permiten determinar la pose (localización y orientación en la imagen) del objeto o residuo, lo que facilita después implementar métodos para llevar a cabo tareas de coger, levantar y soltar con robots manipuladores. Por ejemplo, en \citep{feng2021} se hace uso de una red MobileNet como base para  implementar una R-CNN de segmentación de regiones \citep{minaee2020}, a partir de las características que extrae MobileNet. Así, la nueva MobileNet + Mask-RCNN consigue mantener altos valores de precisión en la detección, superiores al noventa por ciento sobre una RaspBerry Pi, permitiendo además la detección de múltiples instancias de objetos en una misma imagen. Y en \citep{kiyokawa2021} se presenta un sistema robótico basado en un KUKA LBR iiwa 14 R820 con cámara RGBD que implementa un sistema de percepción visual basado en una red neuronal SSD para la localización, reconocimiento, recogida y clasificación de residuos domésticos en entornos de interior. Es en este tipo de aproximaciones en las que se basa la idea de nuestro trabajo. 

En cuanto a la tarea de estimación de poses de agarre, se han aplicado diversos enfoques que pueden clasificarse en dos categorías: basados en datos, como propusieron en \citep{sahbani2012overview} y \citep{liu2021dynamic}, y basados en modelo como hicieron en \citep{jiang2021manipulator} y \citep{shaw2018tactile}.  Los trabajos basados en datos analizan datos muestreados de las tareas de agarre y/o manipulación con técnicas de aprendizaje \citep{newbury2022deep}.
En los basados en modelo, se emplean características del manipulador y se extraen descriptores del objeto a manipular para crear un modelo físico \citep{bohg2013data}.  
Además, es posible calcular la pose para realizar el agarre, tal y como propone \citep{guo2020pose} o el cálculo de los puntos de agarre en los objetos \citep{kim_li_lee_2021}, siendo este último tipo de método el que se empleará.

Con respecto a la manipulación de objetos, se han realizado numerosos trabajos en el estado del arte con el fin de incluir los robots manipuladores en ámbitos como el social o el industrial. Tradicionalmente, se han empleado sensores capaces de medir la fuerza producida por un objeto sobre el extremo del robot durante la manipulación. Estos sensores permiten desarrollar estrategias de control para llevar a cabo tareas complejas de manipulación. Más recientemente, se han desarrollado trabajos con sensores táctiles con diferentes tecnologías como piezo-resistiva, capacitiva, óptica, magnética o barométrica, que devuelven valores de presión ejercidos sobre el objeto manipulado, como por ejemplo \citep{yao2020highly},  \citep{zapata2019learning} y \citep{velasco2020clasificacion}. En \citep{yao2020highly}, los autores diseñan un sensor capacitivo tridimensional para la detección de contacto y deslizamiento obteniendo notables resultados gracias a la alta sensibilidad, bajo tiempo de respuesta y baja histéresis de estos sensores. En \citep{zapata2019learning} se utilizan los sensores BiotacSP acoplados a la mano robótica Shadow con el objetivo de detectar deslizamiento de objetos mediante redes neuronales. Por otro lado, en \citep{velasco2020clasificacion} los autores utilizaban técnicas clásicas de aprendizaje automático para clasificar objetos según los valores articulares de una mano Allegro y los valores de presión de los sensores resistivos colocados sobre los dedos. La tendencia actual se centra más en sensores táctiles de tipo óptico (basados en imagen) como Gelsight \citep{yuan2017gelsight}, GelSlim \citep{donlon2018gelslim}, Tactip \citep{ward2018tactip} o Digit \citep{digit}. A diferencia de los sensores mencionados anteriormente (fuerza/presión), este tipo de sensores no proporcionan valores de fuerza o presión, sino una imagen en color donde se produce el contacto entre el sensor y el objeto. De esta manera, se pueden aplicar las técnicas más novedosas en el campo de visión por computador y deep learning para tareas de manipulación y agarre basadas en imagen. Por ejemplo, en \citep{kolamuri2021improving} realizan un estudio para compensar la rotación indeseada de un objeto a partir de las imágenes táctiles de sensores Gelsight, mediante técnicas tradicionales de visión por computador. Otro ejemplo sería el trabajo de \citep{lin2022tactile} donde aplican técnicas de aprendizaje por refuerzo para llevar a cabo diferentes tareas de manipulación con sensores táctiles como Gelsight, Digit o Tactip. En este trabajo se pretende continuar con la tendencia de uso de los sensores táctiles de bajo coste basados en imagen, ya que aportan más información sobre las características de los objetos que los sensores de fuerza o presión, ya que tienen una resolución espacial mayor.

Las principales novedades de nuestra propuesta son varias.
Primero, se ha mejorado el proceso de reconocimiento, empleando una arquitectura Yolact para segmentación \citep{liu2020}, la cual pertenece a la misma familia de redes que Mask-RCNN o SDD \citep{minaee2020}, para así conseguir más rapidez de detección sin sacrificar tasa de precisión. Después, se mejora el sistema visual incorporando un sistema de estimación de los puntos de agarre a partir de la detección del objeto-basura, extendiendo el trabajo que empezó a desarrollarse en \citep{gea2021}. Y finalmente, se ha incorporado un sistema táctil basado en CNNs y sensores táctiles Digit, para controlar la manipulación durante la tarea de agarrar y levantar los residuos, a partir de desarrollos previos mostrados en \citep{castano2021}. Todo esto se ha empezado a testar, funcionando en situaciones similares a las reales en entornos de exterior.

\section{Plataforma robótica de manipulación}

Nuestro sistema robótico para la recogida de residuos domésticos consta de una plataforma móvil de fabricación propia, conocida como BLUE \citep{pino2020}, similar a otras plataformas de manipulación móvil como la presente en \citep{suarez2020manipulador}. Esta plataforma recientemente ha sido modificada para transportar un manipulador UR5e de 6 grados de libertad, dotado de una pinza de 2 dedos ROBOTIQ-2F140. Mientras la plataforma móvil se emplea para la navegación en busca de residuos, éste último se emplea para las tareas de recogida. 

\begin{figure}[H]
\centering
    \includegraphics[width=8cm]{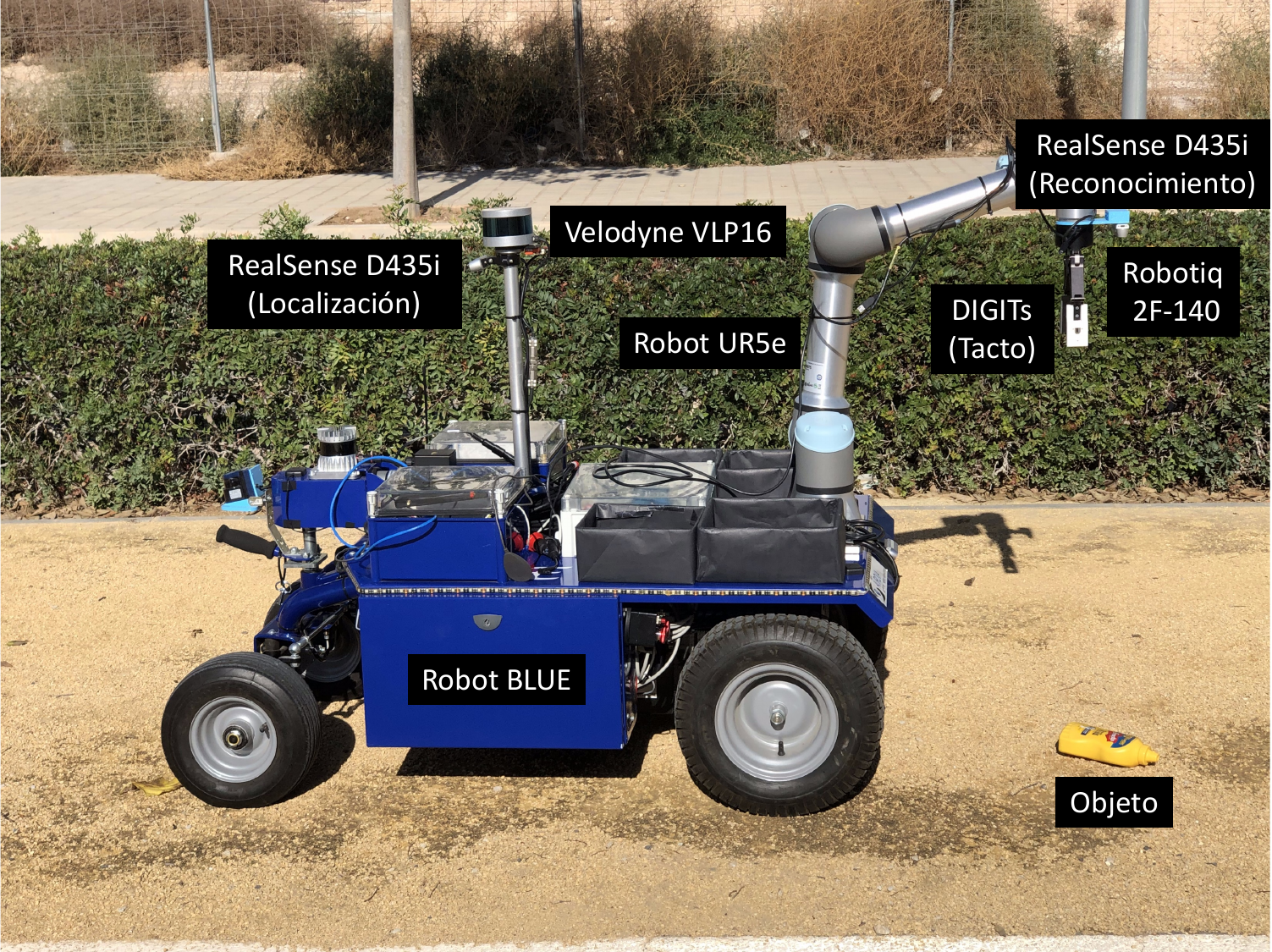}
    \caption{Plataforma robótica y sensores} 
    \label{fig:blue}
\end{figure}

A parte de estos dispositivos robóticos, el sistema consta de un subsistema de percepción visual y otro táctil, ambos constituidos por una serie de sensores externos y por los algoritmos que permiten adquirir datos, procesarlos, e interpretar tanto el escenario como la interacción del robot con el entorno. Los sensores visuales empleados son cámaras RealSense D435i, adquiriendo imágenes RGBD a 640x480 a una velocidad de 30 \gls{fps}. También, se emplea un 3D LiDAR Velodyne VLP16 que trabaja a 10Hz y es capaz de medir distancias hasta 100 m, frente al rango de distancias de uso de $0,3-3$m que proporcionan las cámaras. Por otro lado, el subsistema táctil consta de dos sensores ópticos de bajo coste, DIGIT \citep{digit}, colocados en el interior de los dedos de la pinza. Cada uno de estos sensores captura imágenes RGB con una frecuencia de hasta 30 \gls{fps} y una resolución de 240x320 píxeles. En la Figura~\ref{fig:blue} se muestran la plataforma robótica y los sensores empleados en este trabajo.

\begin{figure}[H]
    \centering
    \includegraphics[width=7cm]{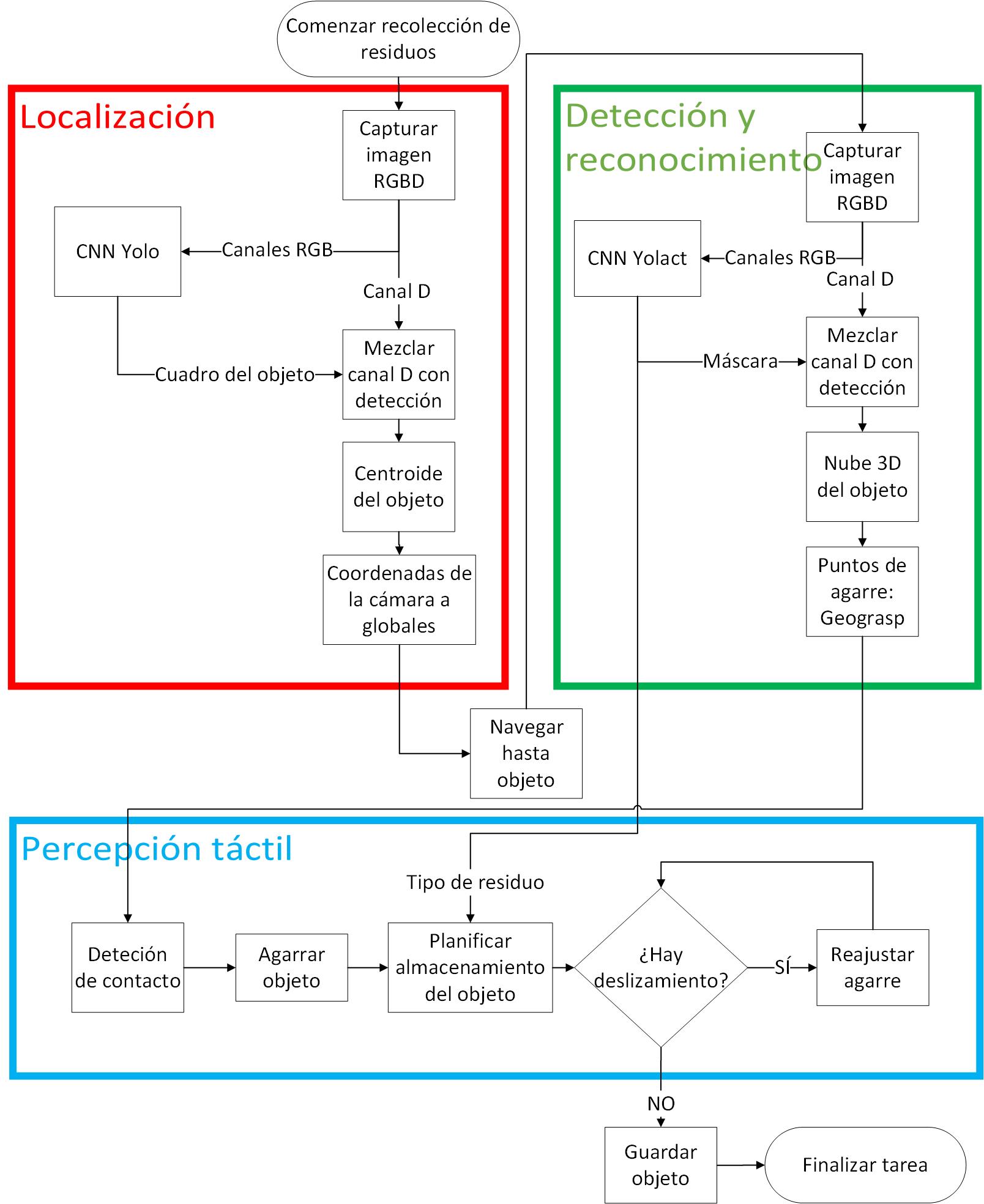}
    \caption{Arquitectura del sistema de percepción visual-táctil para la recolección robótica de residuos} \label{fig:esquema}
\end{figure}

Los dispositivos robóticos, tanto los de desarrollo propio (BLUE), comerciales (UR5e y pinza), como los sensores (RGBD, LiDAR), etc. se han integrado en ROS corriendo sobre Ubuntu Melodic Morenia para facilitar el intercambio de información y la comunicación. Así, sobre ROS se han implementado todos los métodos y algoritmos que forman parte de los módulos software de los subsistemas de perpepción visual y táctil. En concreto, estos métodos y algoritmos se ejecutan en una Jetson AGX Xavier. En la Figura~\ref{fig:esquema} se muestra la arquitectura general del sistema. El sistema presentado está orientado a la recogida de diferente tipología de residuos domésticos no orgánicos y sólidos, como botes, botellas, latas o tetrabricks, en ambientes exteriores, con un manipulador móvil. Éste se encuentra dividido en tres partes diferenciadas, siendo éstas la localización de los residuos y la navegación para su recogida (Sección \ref{sec:localizacion}), la detección de estos residuos una vez estamos cerca de los mismos y el reconocimiento para su categorización (Sección \ref{sec:deteccion}) y finalmente, la manipulación de los mismos para agarrarlos y subirlos a bordo de la plataforma robótica (Sección \ref{sec:manipulacion}).


\begin{figure*}
\centering
\includegraphics[scale =0.24] {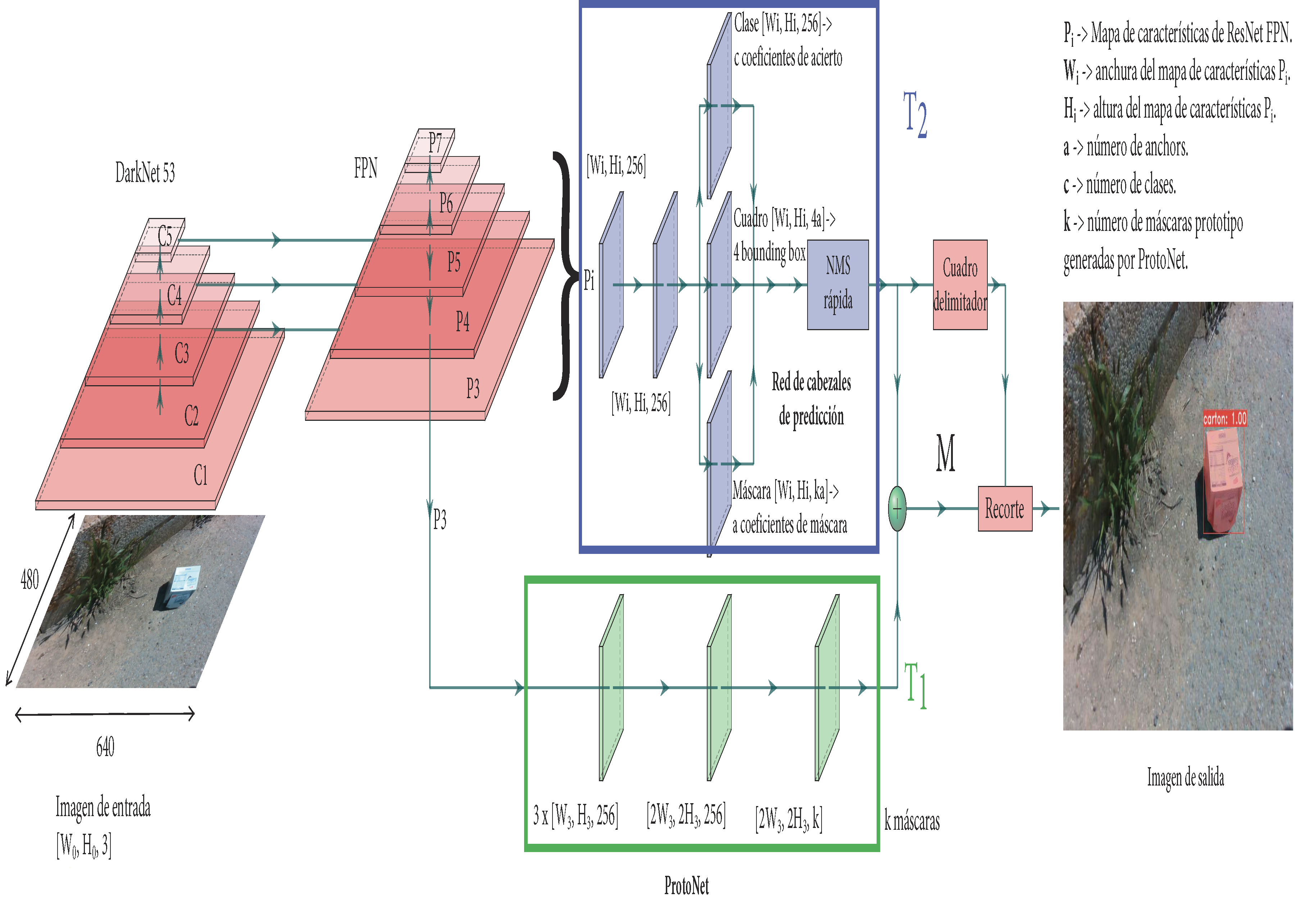}
\caption{Yolact: red neuronal empleada para realizar la segmentación de objetos}
\label{fig:yolact_scheme}
\end{figure*}

\section{Sistema de percepción visual}
\subsection{Detección y reconocimiento en exteriores} 
\label{sec:deteccion}

Una vez se han detectado potenciales residuos a partir de las imágenes de una de las cámaras RGBD y calculado las coordenadas 3D de cada uno de ellos (se detallará en la sección siguiente), tal como se muestra en el procedimiento resaltado en color rojo en la Figura~\ref{fig:esquema}, el robot se aproxima navegando hasta ellos para tratar de reconocer el objeto con mayor exactitud. Este proceso se detalla en verde en la misma Figura~\ref{fig:esquema}. Para ello, se ha implementado un módulo software específico para esta tarea. El módulo hace uso de una \gls{cnn} de tipo Yolact \citep{bolya2019yolact} mostrada en la Figura~\ref{fig:yolact_scheme}. Esta red neuronal fue escogida ya que fue la que mejores resultados obtuvo en comparación con otros métodos en el estado del arte en la tarea de segmentación de objetos como la segmentación binaria o la asistida o semiautomática. Yolact es un detector monoetapa que realiza el proceso de segmentación en un solo paso y en tiempo casi real. Además, la familia Yolact permite emplear distintas arquitecturas de red, tales como ResNet \citep{He_2016_CVPR} con distinto número de capas, DarkNet \citep{darknet13}, etc. En todas esas arquitecturas base se incluye una red de extracción de características, conocida como \gls{fpn} o red piramidal de extracción de características, la cual consiste en variar el tamaño de la imagen para obtener varios niveles de características semánticas.

La detección monocapa $M$ se divide en dos subtareas paralelas $T_1$ y $T_2$, marcadas en verde y azul en la Figura~\ref{fig:yolact_scheme}. A partir de $P_3$ de la \gls{fpn}, $T_1$ crea un conjunto de regiones candidatas en las que pueden existir objetos de interés, generando $k$ máscaras. Y haciendo uso de todas las capas \gls{fpn}, $T_2$  genera $c$ coeficientes de acierto, 4 regresores de $\gls{bb}$ o cuadro delimitador y $a$ coeficientes de máscara, uno por cada prototipo ($4 + c + a$). Una vez finalizado el proceso en paralelo, se fusionan ambas subtareas usando una combinación lineal de la primera tarea $T_1$ y la transpuesta de la segunda $T_2^T$, seguido por un proceso sigmoide no lineal $\sigma$, como se indica en (\ref{eq:m}).

\begin{equation} 
\label{eq:m}
M=\sigma (T_1T_2^T)
\end{equation}

La función de pérdida de esta \glspl{cnn} viene definida por la suma ponderada de tres funciones de pérdida como se indica en (\ref{eq:sum_loss}), donde $L_{cls}$ representa la pérdida de la clase, $L_{box}$ la del cuadro delimitador y $L_{mask}$ la de la máscara.

\begin{equation} 
\label{eq:sum_loss}
L_{yolact} = 1.0\;L_{cls} + 1.5\;L_{box}+6.125\; L_{mask}
\end{equation}

La pérdida de la clase $L_{cls}$, como se muestra en (\ref{eq:l_cls}), representa la suma de las coincidencias positivas y negativas del cuadro delimitador, ambas calculadas con una función softmax.

\begin{equation} 
\label{eq:l_cls}
\begin{split}
L_{cls} = -\left( \frac{1}{N} \right)\;&\sum_{i\in Pos}x_{i,j}^{p}\;log(\widehat{c}_{i}^{p})\;-\;\sum_{i\in Neg}\;log(\widehat{c}_{i}^{0})
\\
& siendo \;\;\; \widehat{c}_{i}^{p} = \frac{e^{c_{i}^{p}}}{\sum_{p}e^{c_{i}^{p}}}
\end{split}
\end{equation}

donde $N$ representa el número de cuadros delimitadores detectados, $x_{i,j}^{p} = \left\{ 1,0\right\}$ indica la coincidencia entre el cuadro delimitador detectado $i$ y el real $j$ para cada categoría $p$, siendo $c_{i}^{p}$ la confianza de las múltiples clases $c$ y $\widehat{c}_{i}^{p}$ la función softmax para cada cuadro delimitador $i$ para cada categoría $p$.

Por otro lado, $L_{box}$ representa la diferencia entre el cuadro delimitador predicho $l$ y el real $g$ cuando hay coincidencia positiva en la detección, y se define como en (\ref{eq:l_box}). Cada cuadro delimitador viene definido por su centro $(cx,cy)$ y dimensiones en anchura y altura $(w,h)$ en píxeles.

\begin{equation} 
\label{eq:l_box}
\begin{split}
L_{box}& = \;\left( \frac{1}{N} \right) \; \sum_{\underset{}{i\;\in \; Pos}}\;\sum_{\underset{}{m\;\in \; \left\{ cx,cy,w,h \right\}}}x_{i,j}^{k}\;smooth_{L1}\left( l_{i}^{m} - \widehat{g}_{j}^{m} \right)
\\
&siendo \;\;\; \widehat{g}_{j}^{cx} = \frac{\left( g_{j}^{cx} - d_{i}^{cx}\right)}{d_{i}^{w}} \;\;\;\;\;\;\;\;
\widehat{g}_{j}^{cy} = \frac{\left( g_{j}^{cy} - d_{i}^{cy}\right)}{d_{i}^{h}}
\\
&\;\;\;\;\;\;\;\;\;\;\;\; \widehat{g}_{j}^{w} = log \left( \frac{g_{j}^{w}}{d_{i}^{w}} \right) \;\;\;\;\;\;\;\;\;\;\;\;
\widehat{g}_{j}^{h} = log \left( \frac{g_{j}^{h}}{d_{i}^{h}} \right)
\\
\end{split}
\end{equation}
 
donde $x_{i,j}^{k}$ vuelve a representar nuevamente la coincidencia entre el cuadro delimitador detectado.

Finalmente, $L_{mask}$ es la pérdida de máscara, que se define como una función de entropía cruzada binaria para cada píxel entre la máscara predicha y la real, tal y como se muestra en (\ref{eq:l_mask}).

\begin{equation} 
\label{eq:l_mask}
L_{mask} = -\left( \frac{1}{N} \right)\;\sum_{i=1}^{N}y_{i}\;log(p(y_{i}))\;+\;(1-y_{i})\;log(1-p(y_{i}))
\end{equation}

donde $y$ representa la categoría, siendo $p(y)$ la probabilidad predicha de tener la categoría.

\begin{figure}[H]
\centering
\subfigure{
    \includegraphics[scale =0.19] {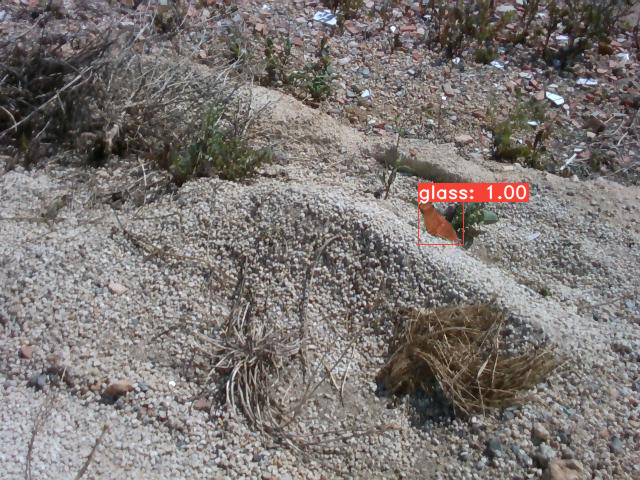}
   \includegraphics[scale =0.19] {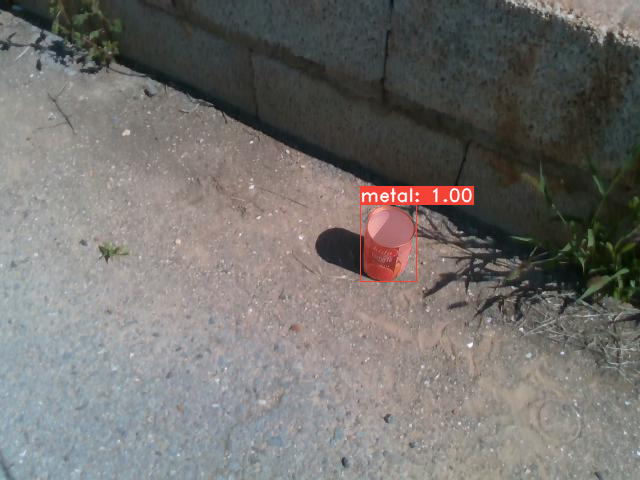}
   \label{subfig1}
 }
 \subfigure{
   \includegraphics[scale =0.19] {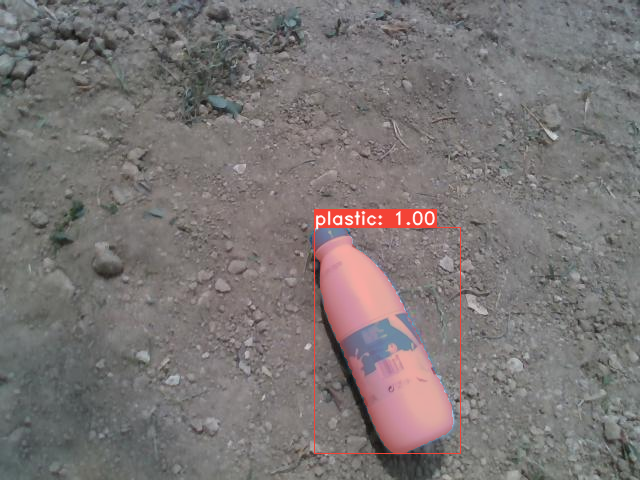}
   \includegraphics[scale =0.19] {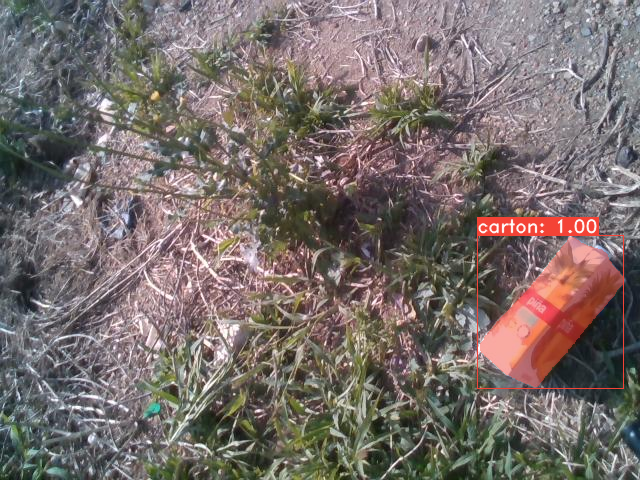}
   
   \label{subfig2}
 }
\caption{Ejemplos de reconocimiento de basura doméstica (botellas, tetrabricks, latas) en entornos de exterior}
\label{fig:objects_yolact}
\end{figure}

Tras realizar un análisis de porcentajes de acierto y tiempos de ejecución, se decidió emplear Yolact con la arquitectura DarkNet de 53 capas. Esta arquitectura fue entrenada durante 30000 iteraciones empleando SGD (Descenso por gradiente estocástico), comenzando con una tasa de aprendizaje de 0.001, un decaimiento del peso de 0.0005 y un impulso de 0.9.  Con ella se consiguió la mejor tasa de reconocimiento, medida en términos de precisión media, $mAP$, además de usar menor tiempo de inferencia. En la Figura~\ref{fig:objects_yolact} se muestran varios ejemplos de reconocimiento de basura doméstica en distintas condiciones de luz y entornos.

\begin{figure}[H]
\centerline{\includegraphics[width=.47\textwidth]{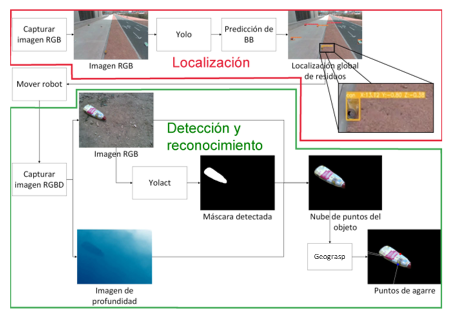}}
\caption{Esquema del proceso de localización, reconocimiento y estimación de puntos de agarre del residuo.}  \label{fig:geograsp}
\end{figure}

Después de reconocer con exactitud el tipo de residuo doméstico, el siguiente paso consiste en estimar por dónde llevar a cabo el agarre para recogerlo. Para hacerlo, se ha  optado por emplear el método propio conocido como Geograsp \citep{zapata2019}, que se basa en la detección de puntos de agarre candidatos empleando nubes de puntos, pero en esta ocasión acometiendo variaciones como las propuestas previamente en \citep{gea2021}.
En concreto, tras realizar con la cámara RealSense D435i la captura de la imagen RGBD, se obtiene la nube de puntos $N = \{p_1,p_2,...,p_m\}$, formada por los puntos $p_i=\{x,y,z\}$ con colores $r$, $g$, $b$ y siendo $i \in \mathbb{N}$, $1 < i < m$. Tras ello, se analiza con Yolact (ver Figura \ref{fig:yolact_scheme}) obteniendo la máscara del objeto. Al tener dicha máscara se escogen los píxeles de la imagen de profundidad en los que se encuentra el objeto, obteniendo una nube de puntos tridimensional a color $N_p = \{p_1,p_2,...,p_n\} \subseteq N$, siendo $m>n$. Esta nube de puntos se envía al algoritmo GeoGrasp, obteniendo sobre ella los puntos de agarre del objeto $G = \{p_a,p_b\} \subseteq N_p$ que se empleará para manipular el objeto. Estos puntos $p_a$ y $p_b$ cumplen las siguientes condiciones: se encuentran más cerca del plano de grasping, poseen la menor curvatura, son opuestos y no son paralelos con el plano de corte. Todo este procedimiento puede verse en la parte inferior de la Figura~ \ref{fig:geograsp}.


\subsection{Localización y estimación de agarre}  
\label{sec:localizacion}

Para llevar a cabo la recogida de residuos, es necesario determinar las coordenadas de localización de todos ellos respecto a la plataforma móvil y, después determinar su ubicación en el mundo con respecto al UR5e que transporta. Por un lado, para poder determinar su ubicación y poder así navegar hasta ellos, y por otro lado, para determinar las coordenadas sobre la superficie de éstos para llevar a cabo el agarre y recogida. 

\begin{figure*}
\centering
\includegraphics[scale =0.32] {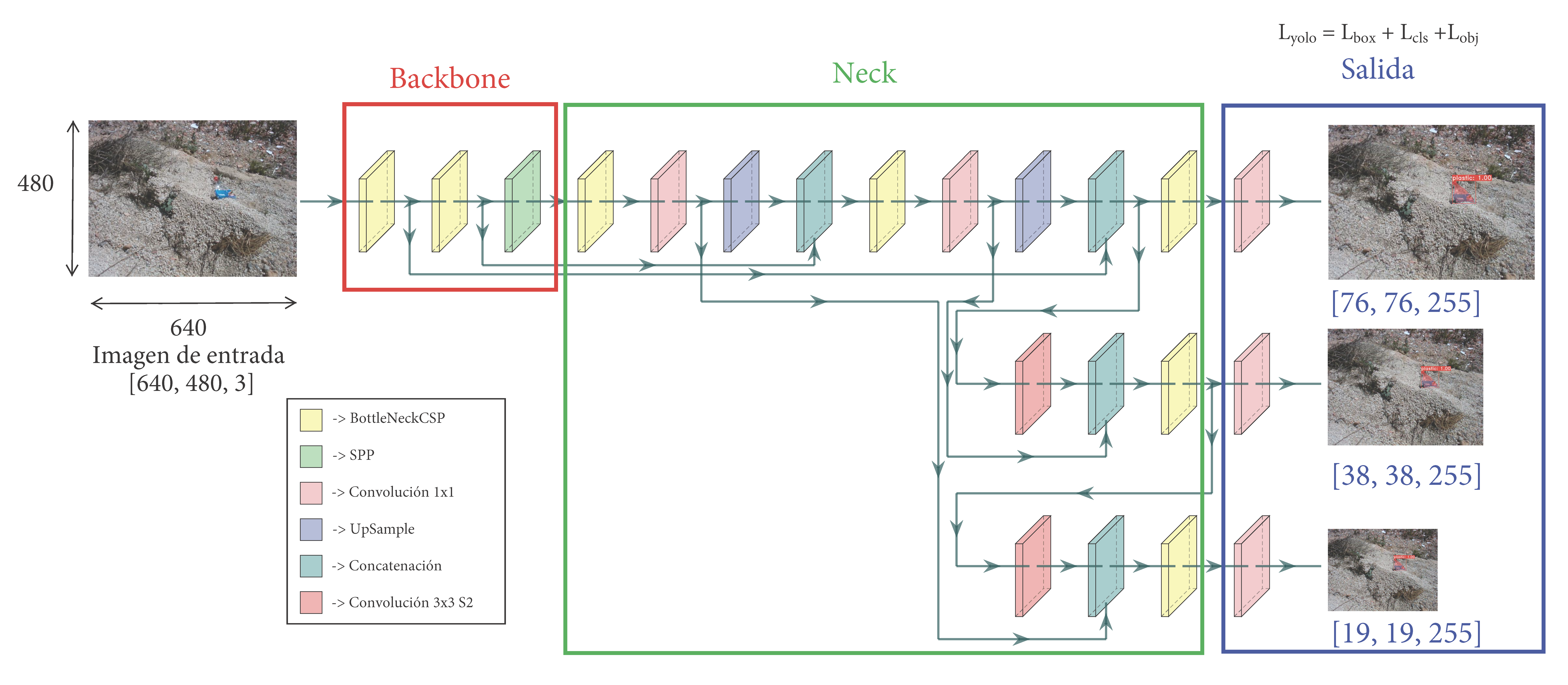}
\caption{Yolo: red neuronal empleada para realizar la localización de objetos}
\label{fig:yolo_scheme}
\end{figure*}

Para el primer paso, se emplea una \gls{cnn} de tipo Yolo. La selección de esta arquitectura de red está fundamentada en el rendimiento que ofrece a nivel de detección de objetos frente a otras del estado del arte como los métodos de dos etapas. Su arquitectura se muestra en la Figura \ref{fig:yolo_scheme} y se encuentra formada por tres partes diferenciadas. El backbone (en rojo) es usado para extraer características importantes que den información a partir de la imagen de entrada dada. El neck (en verde) se emplea para generar pirámides de características. Éstas ayudan a generalizar la escala, permitiendo identificar el mismo objeto en imágenes de distinto tamaño. Por último, la salida (en azul) proporciona el resultado que se obtiene tras procesar la imagen por las capas anteriores.
La función de pérdida de esta \gls{cnn} viene dada por el sumatorio de tres funciones de pérdida (\ref{eq:loss}):

\begin{equation} 
\label{eq:loss}
L_{yolo} = L_{box}+L_{cls}+L_{obj}
\end{equation}

donde $L_{box}$ representa la pérdida del cuadro delimitador, $L_{cls}$ la de la clase y
$L_{obj}$ la de la confianza o seguridad de la detección, que se estiman como  (\ref{eq:loss_box}), (\ref{eq:loss_cls}) y (\ref{eq:loss_obj}) respectivamente.

\begin{equation} 
\label{eq:loss_box}
\begin{split}
L_{box} = &\lambda_{coord}\sum_{\overset{}{i=0}}^{S^{2}}\sum_{\overset{}{j=0}}^{B}I_{i,j}^{obj}(2-w_{i}×h_{i}) \;\Bigg[\Bigg.\left(x_{i}-\widehat{x}_{i^{}}^{j}\right)^{2} + \\
&+\left(y_{i}-\widehat{y}_{i^{}}^{j}\right)^{2} +\left(w_{i}-\widehat{w}_{i^{}}^{j}\right)^{2} + \left(h_{i}-\widehat{h}_{i^{}}^{j}\right)^{2}\Bigg. \Bigg]
\end{split}
\end{equation}

Los términos $\lambda_{coord}$ representan los coeficientes de pérdida de posición (peso que tiene respecto a los otros elementos en (\ref{eq:loss})), siendo $S$ cada una de las porciones de búsqueda en las que se divide la imagen de entrada, $B$ los anchors o cuadro delimitador predefinido de cierta altura y anchura para cada $S$ e $I_{i,j}^{obj}$ el cuadro delimitador en $(i,j)$ que contiene al objetivo, que toma el valor 1 si lo contiene o 0 en caso contrario.

Además, la región del cuadro delimitador está definida como ($\widehat{x}$, $\widehat{y}$, $\widehat{w}$, $\widehat{h}$) que representan las coordenadas del centro, así como su anchura y altura; mientras que ($x$, $y$, $w$, $h$) son los valores estimados para el cuadro delimitador por la \gls{cnn} en cada búsqueda.

Por otro lado, $L_{cls}$ y $L_{obj}$ se calculan como (\ref{eq:loss_cls}) y (\ref{eq:loss_obj}). 

\begin{equation} 
\label{eq:loss_cls}
L_{cls} = \lambda_{class}\sum_{\overset{}{i=0}}^{S^{2}}\sum_{\overset{}{j=0}}^{B}I_{i,j}^{obj}\sum_{\overset{}{c\;\in\; clases}}p_{i}\left( c \right)\;\widehat{p_{l}}\left( c \right)
\end{equation}

\begin{equation} 
\label{eq:loss_obj}
\begin{split}
L_{obj} &= \lambda_{noobj}\sum_{\overset{}{i=0}}^{S^{2}}\sum_{\overset{}{j=0}}^{B}I_{i,j}^{noobj} \left(c_{i} -\widehat{c}_{l} \right)^{2} + \\
&+\lambda_{obj} \sum_{\overset{}{i=0}}^{S^{2}} \sum_{\overset{}{j=0}}^{B}I_{i,j}^{obj} \left(c_{i} -\widehat{c}_{l} \right)^{2}
\end{split}
\end{equation}

donde $\lambda_{class}$ son los coeficientes de pérdida de categoría, $c$ el número total de clases, $p_{i}\left(c \right)$ la probabilidad de la categoría del elemento analizado y $\widehat{p_{l}}\left(c \right)$ es el valor real de la categoría. Siendo, además, $\lambda_{noobj}$ y $\lambda_{obj}$ los pesos de la confianza cuando no existe y existe un objeto en el cuadro delimitador y $c_{i}$, $\widehat{c_{l}}$ las confianzas real y predicha del cuadro delimitador.

La salida de la \gls{cnn}, por lo tanto, proporciona un cuadro delimitador en la imagen RGBD de entrada, según el procedimiento de localización mostrado en la Figura~\ref{fig:esquema}. Después, se extraen las coordenadas en píxeles $(u,v)$ del centro del cuadro delimitador del objeto-basura, y se transforman a coordenadas 3D respecto de la cámara con la que se obtuvo su detección. 

Para ello, se emplean los parámetros intrínsecos  $K$ de la matriz de calibración de la cámara y la distancia $d$ obtenida a partir del canal de profundidad $D$, operando tal y como se indica en (\ref{eq:cam_real}).

\begin{equation} 
\label{eq:cam_real}
(x_c,y_c,z_c)=K \cdot (d,u,v) 
\end{equation}

Después, se transforman las coordenadas obtenidas a coordenadas con respecto al LiDAR como se expresa en (\ref{eq:real_lidar}).

\begin{equation} 
\label{eq:real_lidar}
\begin{bmatrix}
x_l \\ y_l \\ z_l \\ 1
\end{bmatrix}
= T_{cl}^l 
\begin{bmatrix}
x_c \\ y_c \\ z_c \\ 1
\end{bmatrix}
\end{equation}

donde $T_{cl}^l$ representa la transformación entre la cámara y el LiDAR. Una vez obtenidas las coordenadas respecto del LiDAR se transforman a coordenadas respecto de la plataforma móvil mediante la transformación fija ($T_l^r$) como se indica en (\ref{eq:lidar_mundo}). Esta transformación refleja la relación entre la ubicación del LiDAR y el sistema GPS de la plataforma móvil. Además, teniendo en cuenta la posición de la plataforma móvil, según el GPS y la odometría, se obtiene un \textit{offset} como la diferencia entre la posición de partida y su posición actual. Así, es posible referenciar con coordenadas relativas a la zona de trabajo actual en vez de con coordenadas GPS globales.

\begin{equation} 
\label{eq:lidar_mundo}
\begin{bmatrix}
x_m \\ y_m \\ z_m \\ 1
\end{bmatrix}
= T_l^r
\begin{bmatrix}
x_l \\ y_l \\ z_l \\ 1
\end{bmatrix}
+ {offset}
\end{equation}

 Un ejemplo del resultado de este procedimiento se puede observar en la parte superior de la Figura~\ref{fig:geograsp}. En él se muestra la detección y localización de varias instancias de dos tipos de objetos distintos en una misma vista, así como el detalle de las coordenadas obtenidas para uno de ellos con respecto a la plataforma móvil.

En el segundo paso, una vez que el robot BLUE está cerca, y dentro del alcance del manipulador UR5e, es necesario calcular la pose con respecto a éste para llevar a cabo el agarre. Para ello, se procede como se comentó previamente en la Sección \ref{sec:deteccion}. Es decir, se hace uso de Yolact para reconocer con mayor exactitud el tipo de objeto. Después, 
se predicen los puntos de agarre $(p_a,p_b)$ sobre su superficie como se comentó anteriormente y muestra la parte inferior de la Figura~\ref{fig:geograsp}. Por lo tanto, para este caso en vez de utilizar las coordenadas del centro del cuadro delimitador del objeto, se usan las coordenadas imagen de esos dos puntos de agarre $(p_a,p_b)$. Para cada uno de estos dos puntos se transforman del mismo modo a como se procedió para el centro del cuadro delimitador. Esto se hace así, porque como herramienta de agarre se emplea una pinza, y el agarre por un único punto como el centroide solo sería factible si en el efector del UR5e llevara una ventosa. 

\begin{figure}[H]
    \centering
     \begin{subfigure}
     \centering
         \includegraphics[width=0.30\textwidth]{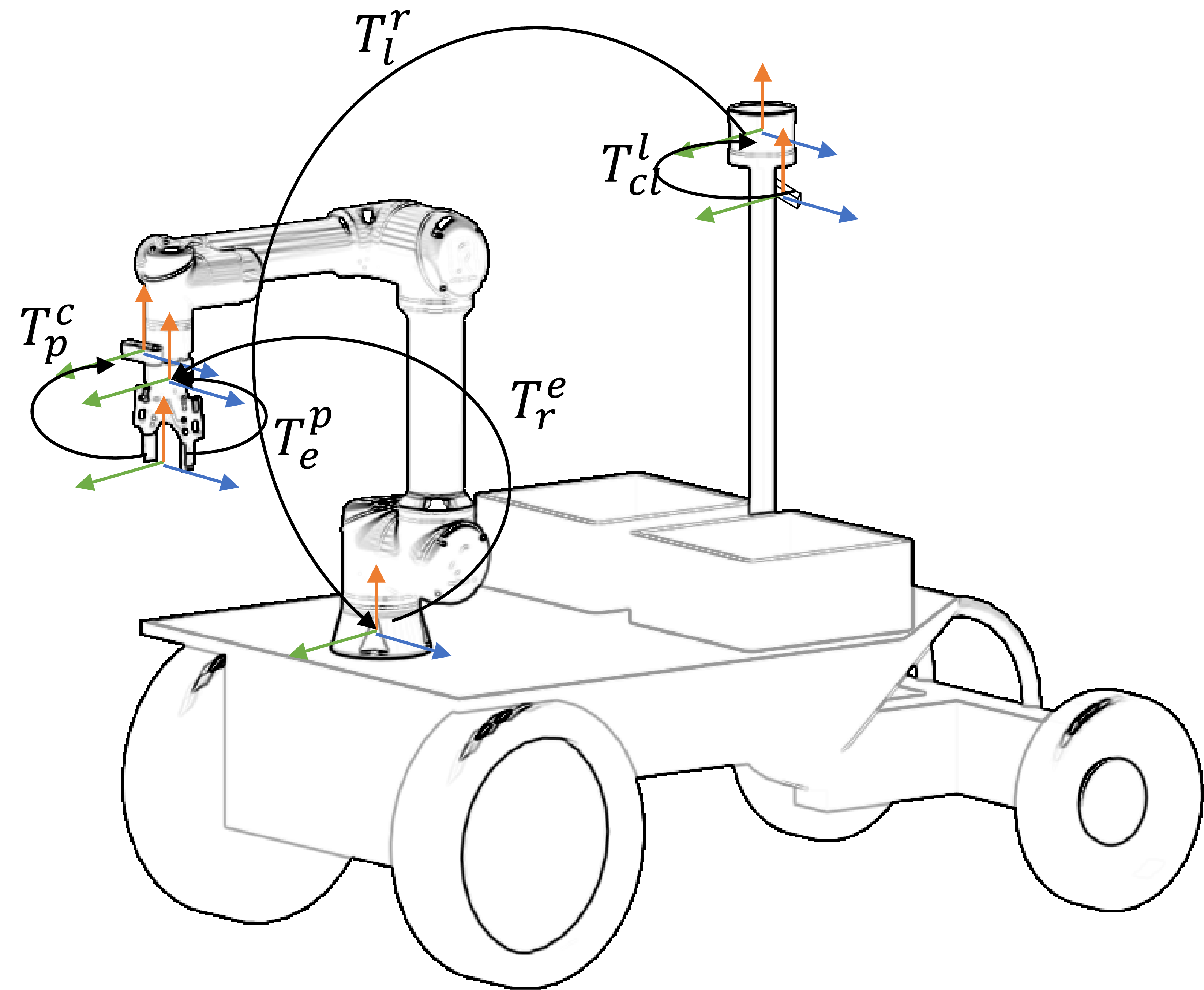}
     \end{subfigure}
     \caption{Configuración de los sistemas de coordenadas a bordo de BLUE en la tarea de navegación y estimación del agarre
     \label{fig:esquema_blue_sistemas} }
\end{figure}

Finalmente, una vez computados $(p_a,p_b)$, es posible determinar la posición 3D de dichos puntos de agarre con respecto al UR5e, como $T=T_{r}^{p} \cdot T_{p}^{c}$. Ya que es conocida la transformación entre la pinza y la cámara situada en el extremo $T_{p}^{c}$ y además, se puede obtener la transformación entre el UR5e y la pinza como $T_{r}^{p}=T_{r}^{e} \cdot T_{e}^{p}$, donde $T_{e}^{p}$ es la transformación del efector a los dedos de la pinza y $T_{r}^{e}$ de la base del robot al efector obtenida mediante el método de Denavit-Hartenberg. Un esquema de la posición de cada uno de los sistemas de coordenadas y transformaciones descritas se puede observar en la Figura \ref{fig:esquema_blue_sistemas}.

\section{Sistema de percepción táctil}
\label{sec:manipulacion}

Para llevar a cabo la tarea de manipulación, que consiste en coger un objeto y dejarlo en una posición final, el robot debe combinar diferentes acciones. Primero, se deberá agarrar el objeto a partir de los puntos de agarre y las trayectorias calculadas previamente. Una vez se ha agarrado el objeto, este es levantado y llevado a la posición de destino donde será depositado. Antes de levantar el objeto, se debe garantizar que su agarre es estable. Después, el robot podrá transportar el objeto reaccionando ante cualquier movimiento indeseado entre el objeto y la pinza robótica. Ambas operaciones, el agarre del objeto y su levantamiento para transportarlo hasta su almacén de destino, deben ser controladas en tiempo real para garantizar una manipulación estable y exitosa (ver proceso en azul en la Figura~\ref{fig:esquema}). Para ello, se ha hecho uso de información táctil obtenida a partir de un sensor óptico de bajo coste, como DIGIT \citep{digit} que se ha incorporado como parte de un control táctil realimentado. Estos sensores fueron acoplados a los dedos de la pinza ROBOTIQ-2F140 del robot UR5e. Al no requerir una manipulación diestra para agarrar los residuos, la forma más económica, sencilla y eficaz de realizar esta manipulación es mediante una pinza robótica de dos dedos.
El subsistema de percepción táctil para controlar el agarre robótico se ha diseñado e implementado para adaptar la tarea de agarre en función de dos algoritmos de detección táctil: uno para la detección de contacto entre objeto y dedos de la pinza, y otro para la detección de deslizamiento que evite que el objeto resbale y caiga. 
El controlador de agarre hace uso del algoritmo propuesto para la detección de contacto y de las imágenes extraídas de los sensores táctiles. Esta tarea se formula como una clasificación de imagen de tipo binario, donde el resultado de la predicción será $0$ o $1$, donde la etiqueta 0 significa que no existe contacto, y 1 significa que sí existe contacto. Para realizar esta predicción se utilizan \glspl{cnn} debido a su alta capacidad para extraer y aprender características a partir de imágenes. En \citep{castano2021}, ya realizamos una extensa experimentación entrenando diferentes arquitecturas de \glspl{cnn} para esta tarea, tales como las basadas en la familia VGG \citep{simonyan2014very} e Inception \citep{szegedy2016rethinking} u otras de arquitectura más ligera y portable como MobileNet \cite{sandler2018mobilenetv2}. Los resultados mostraron que la arquitectura InceptionV3 obtenía un tiempo de inferencia 50 ms menor que otras, manteniendo los mismos porcentajes de acierto. Por lo que, aquí se ha optado por utilizar esta arquitectura pero, esta vez, aplicándole una serie de modificaciones para acelerar el tiempo de inferencia para trabajar embebida en la Jetson AGX Xavier. En primer lugar, se extraen características de la capa "mixed5", reduciendo así el número de filtros y parámetros. En segundo lugar, se añaden a continuación las siguientes capas neuronales: una capa de agrupación de promedio global, dos bloques formados por una capa de normalización por lotes, una capa totalmente conexa y una técnica de regularización conocida como "dropout". Finalmente, se añade una última capa totalmente conexa con una neurona y una función de activación sigmoide que devuelve un valor entre 0 y 1. De forma empírica se establece un umbral para redondear estos valores a 0 (no contacto) o 1 (contacto). La función de pérdida utilizada es la entropía cruzada binaria, que ya se definió en (\ref{eq:l_mask}).

\begin{figure}[H]
\centerline{\includegraphics[width=.45\textwidth]{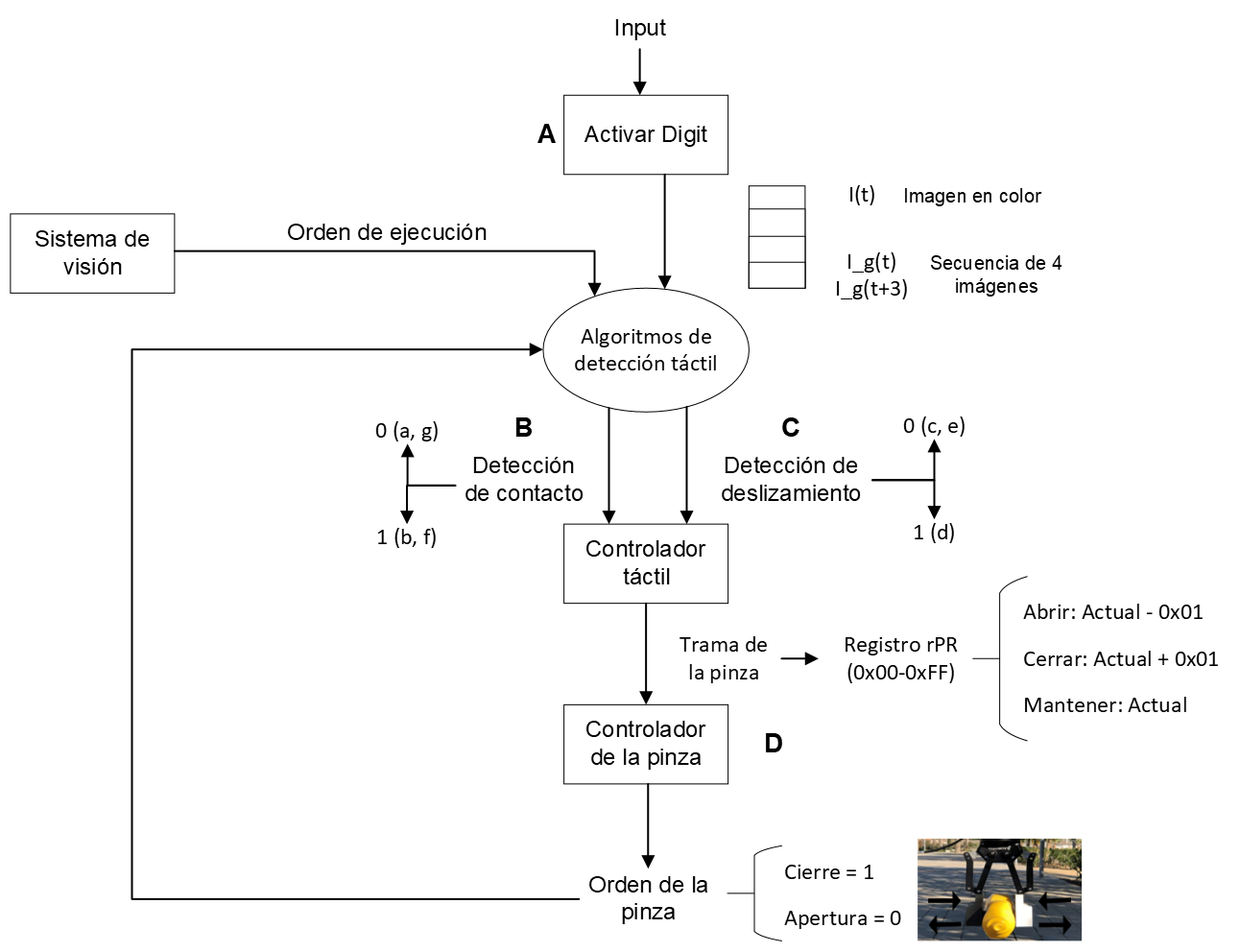}}
\caption{Esquema del controlador de agarre para conseguir estabilidad durante la manipulación. a, b, c, d, e, f y g hacen referencia a la Figura \ref{fig:secuencia_tactil}}  \label{fig:tactil_contacto_slip}
\end{figure}

Como se puede observar en la Figura \ref{fig:tactil_contacto_slip}, tras activarse los sensores táctiles (ver letra A en Figura \ref{fig:tactil_contacto_slip}), el controlador recibe una orden del sistema de visión, indicando si se tiene que agarrar o soltar el  objeto. En primer lugar, se deberá agarrar el objeto, por lo que el sistema utilizará la imagen en color de los sensores. Para el control de agarre se ejecuta el algoritmo de detección de contacto (ver letra B en Figura \ref{fig:tactil_contacto_slip}) mencionado en el párrafo anterior, que enviará la orden de cerrar mediante un registro al controlador de la pinza (ver letra D en Figura \ref{fig:tactil_contacto_slip}) hasta que el algoritmo detecte que el objeto ya se encuentra agarrado. El propio controlador, internamente, cambiará el modo de detección de contacto a deslizamiento mientras el robot levanta y traslada el objeto. Durante este proceso se pueden producir movimientos indeseados como, por ejemplo, deslizamientos que hagan que el objeto resbale. El robot debe ser capaz de ajustar la apertura de la pinza para evitar una caída del objeto no deseada. Para esta, se propone un segundo algoritmo de detección de deslizamientos (ver letra C en Figura \ref{fig:tactil_contacto_slip}). Este algoritmo, también, se formula como una clasificación binaria, donde la predicción con valor 0 corresponde a no deslizamiento, y con valor 1 a deslizamiento. Sin embargo, la metodología es distinta con respecto a la detección de contacto. En este caso no se utilizan las imágenes en color de los sensores DIGIT, sino que se aplica un proceso de preprocesamiento. En primer lugar, se almacena una secuencia temporal de 4 imágenes, en escala de gris, consecutivas en el tiempo. Este número de imágenes es menor que el rango de muestreo requerido en otros sensores dinámicos unidimensionales, como en \citep{s22176456}, porque los sensores táctiles basados en imagen proporcionan información espacial. De este modo, se compensa la carencia de información temporal con una mayor información espacial de cada muestra. A continuación, se realiza una operación de resta entre la última y la primera imagen de la secuencia, $I_{gray}^{t+3}$ y $I_{gray}^t$, obteniendo una imagen binarizada que recoge el estado de la deformación del sensor, $I_{subt}$. El último paso consiste en aplicar una operación morfológica de apertura para eliminar ruido indeseado de la imagen. El resultado es una imagen binarizada $I_{bin}$ tal como se indica en (\ref{eq:img_slip}). Tras el proceso de detección de deslizamiento, una vez el robot se encuentra en la posición de dejada del objeto, el módulo de visión indicará al módulo táctil que debe soltar el objeto. Entonces, el módulo táctil volverá a cambiar de modo de detección a detección de contacto para soltar el objeto, y mover el robot a la posición inicial una vez no se detecte contacto entre el objeto y los dedos de la pinza.

\begin{equation} 
\label{eq:img_slip}
I_{bin} = I_{subt} \circ \kappa = (I_{gray}^{t+3} - I_{gray}^{t}) \circ \kappa
\end{equation}

donde $\kappa$ es el elemento estructurante aplicado en la operación morfológica, cuyo tamaño es de 7x7 y tiene forma de cruz. Este elemento estructurante se ha escogido de forma empírica.

Después, a partir de $I_{bin}$, se calcula el brillo como se indica en (\ref{eq:brillo_slip}), donde $h$ y $w$ representan la altura y anchura de la imagen del sensor en píxeles. 

\begin{equation} 
\label{eq:brillo_slip}
B_{I_{bin}} = \frac{\sum_{i=0}^{h-1} \sum_{j=0}^{w-1} I_{bin} (i,j)}{h \times w}
\end{equation}

Si se ha producido alguna deformación en la secuencia de imágenes táctiles, aparecerá un patrón blanco en la imagen preprocesada, mientras que esta imagen será completamente negra si no se produce ningún deslizamiento. Por lo tanto, se establece un umbral empíricamente para clasificar esta imagen binarizada en el valor 0 (no hay deslizamiento) o 1 (sí hay deslizamiento). Finalmente, el controlador de deslizamiento utiliza este sistema de detección para cerrar la pinza en el caso de detectar un deslizamiento, asegurando así la estabilidad del agarre.

\section{Resultados experimentales} 

\subsection{Rendimiento de los métodos propuestos}

El entrenamiento y validación de las redes neuronales propuestas, se ha hecho con una GPU-NVIDIA A100 con 40 GB de memoria RAM, mientras que para la ejecución en tiempo real se utiliza el sistema embedido NVIDIA Jetson AGX Xavier. En cuanto a las herramientas software, se ha empleado el sistema Docker para mantener simultáneamente diversos contenedores con diferentes versiones de librerías. En el módulo de localización se ha empleado python 3.8, torch 1.7.0 con soporte para CUDA 11.0 y OpenCV 4.5.5.64, mientras que en el módulo de detección se ha empleado python 3.6 y torch 1.7.0.

Con el objetivo de entrenar y validar los módulos de visión y táctil, se ha creado un dataset.
Se han utilizado objetos de diferente material (plástico, cartón, vidrio y metal), ya que estos cuatro son representativos de los residuos urbanos, y uno de los objetivos principales de nuestro sistema es la clasificación de estos materiales. Los diferentes objetos se han colocado en entornos exteriores variados (cemento, tierra, arena, césped, etc) con el objetivo de obtener imágenes con la mayor diversidad posible. Además, se ha creado un dataset variado en cuanto a características referentes al tacto, ya que los objetos difieren según la forma, peso, tamaño, rigidez, textura y fricción. En total se han empleado más de 6900 muestras visuales de objetos en distintos entornos, posiciones y con diferentes condiciones lumínicas de más de 50 objetos. El número de muestras están balanceadas por tipo de material del objeto. 

\begin{figure}[H]
\centerline{\includegraphics[width=.45\textwidth, height=7cm]{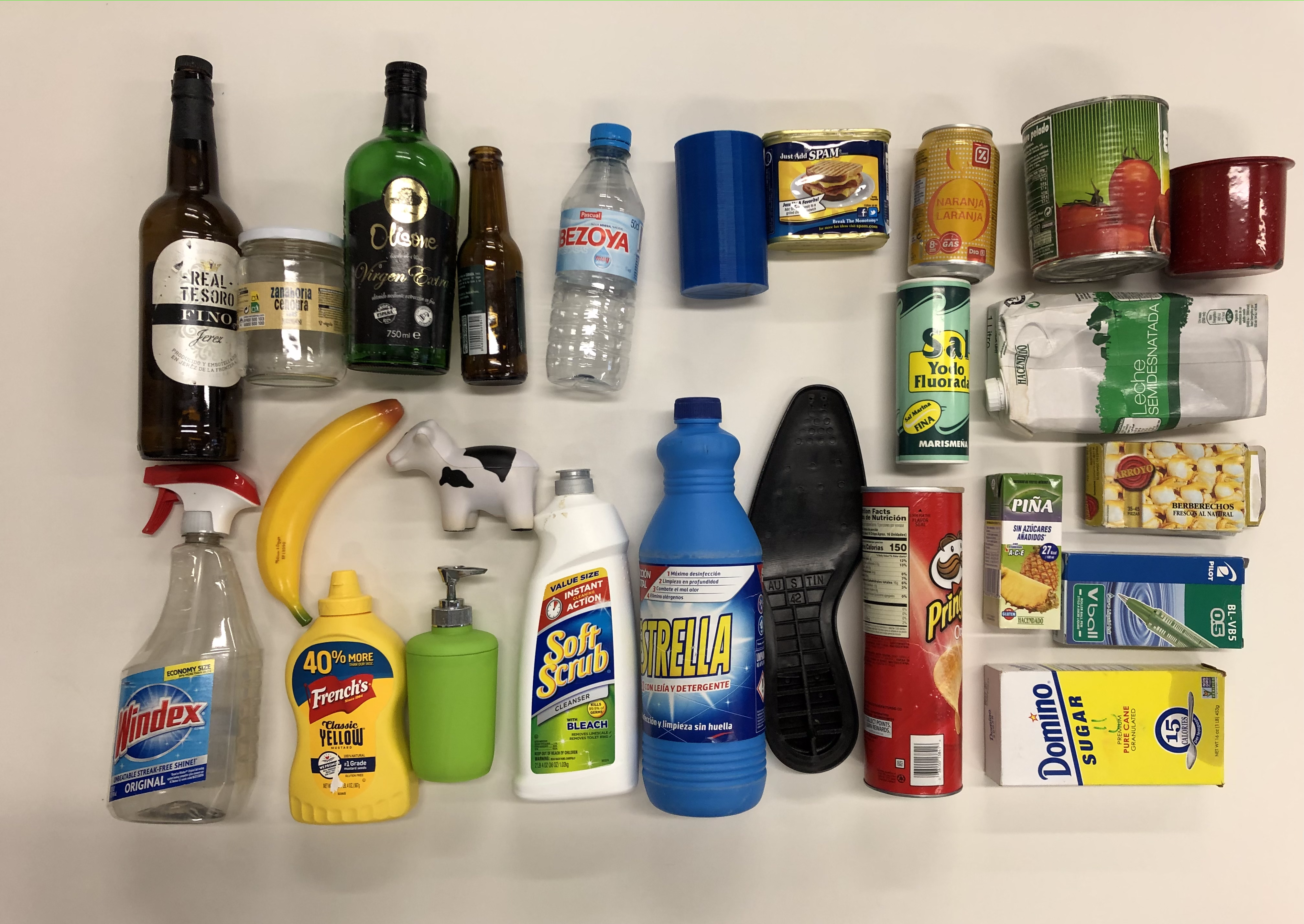}}
\caption{Objetos utilizados para entrenar y validar el sistema visual-táctil}  \label{fig:vision_tactil_dataset}
\end{figure}

La Figura \ref{fig:vision_tactil_dataset} recoge una muestra representativa de los diferentes objetos utilizados para entrenar y validar el sistema visual-táctil. Para el sistema de percepción visual se empleó todo el dataset con la siguiente división aleatoria $70\%, 20\%, 10\%$ para entrenamiento, validación y prueba offline, asegurando muestras distintas en cada subconjunto. Sin embargo para el sistema de percepción táctil se usó un número más reducido de objetos. Concretamente, se emplearon 11 de los 50 objetos con un total de 24607 muestras táctiles de contacto y 3540 muestras táctiles de secuencias de deslizamiento (cada una consta de 4 imágenes táctiles). La elección se realizó atendiendo a características de geometría, dureza y fricción del material de estos, para tener un subconjunto de objetos con comportamientos distintos desde el punto de vista del sentido del tacto. En este caso, la división para entrenamiento, validación y prueba, no se llevo a cabo separando muestras sino utilizando para entrenar el $72\%$ de los objetos y para evaluar el $28\%$ restante. Los conjuntos disjuntos de objetos de entrenamiento y prueba, permiten comprobar la capacidad de genericidad de los algoritmos.

Una vez entrenada la \gls{cnn} mostrada en Figura~\ref{fig:yolact_scheme} (módulo de visión), se ha evaluado su rendimiento en las tareas que implica la actividad de agarre robótico. Para medir los resultados de la detección y reconocimiento visual, se utiliza la métrica de Average Precision (AP) o precisión media como en (\ref{eqn:ap}), de todas las categorías de objetos, haciendo uso de umbrales de confianza de Intersection over Union (IoU) o intersección sobre unión, fijados en 0.5, 0.75 y 0.90. 
En concreto, AP se estima como el área bajo la curva Precision-Recall (PR-AUC) que queda segmentada en $i$ regiones, calculando el Recall y la Precisión para cada región dada como en (\ref{eqn:pr}). 

\begin{equation}
    AP_{IoU} = \sum_{i \in \tilde{R}} (R_{i+1} - R_{i}) \cdot \max_{\tilde{R} \geq R_{i+1}} P(\tilde{R})
\label{eqn:ap}
\end{equation}    

\begin{equation}
R=\frac{TP}{TP+FP} \hspace{1cm}  P=\frac{TP}{TP+FN} 
\label{eqn:pr}
\end{equation} 

El comportamiento obtenido en cuanto precisión para el módulo de reconocimiento visual es:  $AP_{50}=0.998$, $AP_{75}=0.980$ y $AP_{90}=0.696$. Además, el error medio obtenido por el módulo de localización es de $0.178 \pm 0.06$ m. para objetos situados a distancias inferiores a 3 m.

Del mismo modo, la arquitectura InceptionV3 empleada en el módulo de percepción táctil ha permitido alcanzar valores de rendimiento de $A=0.98$ y $A=0.91$ en las tareas de detección de agarre y deslizamiento, respectivamente. Para el módulo táctil se ha escogido la métrica tradicional de precisión (Accuracy) (\ref{eqn:acc}).

\begin{equation}
    A = \frac{TP+TN}{TP+TN+FP+FN}
\label{eqn:acc}
\end{equation}    

donde $TP$ denota que el contacto o deslizamiento detectado es correcto, $TN$ que el sistema está detectando ausencia de contacto o deslizamiento correctamente, mientras que $FP$ y $FN$ indican incorrectas detecciones, es decir se detecta el evento contrario al que realmente se produce.  

Con estos resultados demostramos que los métodos implementados de
nuestro sistema visual-táctil son adecuados para las tareas de recogida de residuos en exteriores que implican localización, reconocimiento y manipulación estable. 

\subsection{Evaluación conjunta del sistema visual-táctil}

\begin{figure}[H]
    \centering
     \begin{subfigure}
     \centering
         \includegraphics[width=0.23\textwidth]{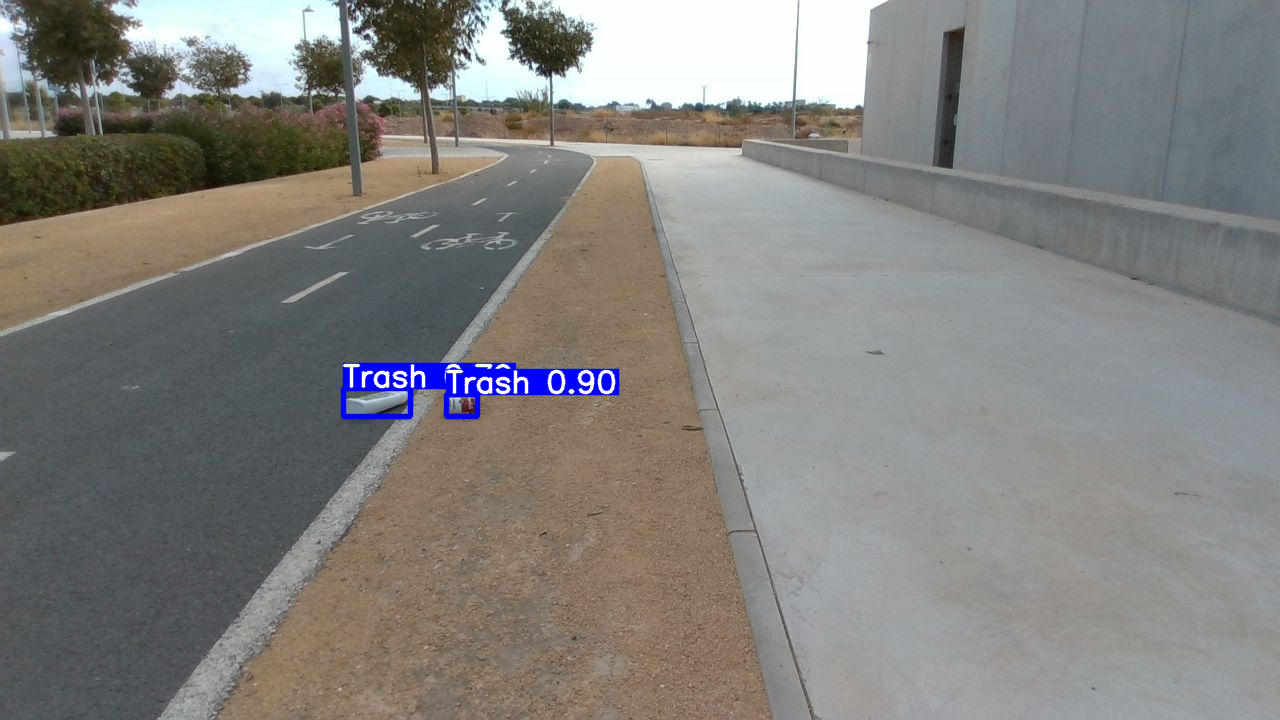}
     \end{subfigure}
     \begin{subfigure}
     \centering
         \includegraphics[width=0.23\textwidth]{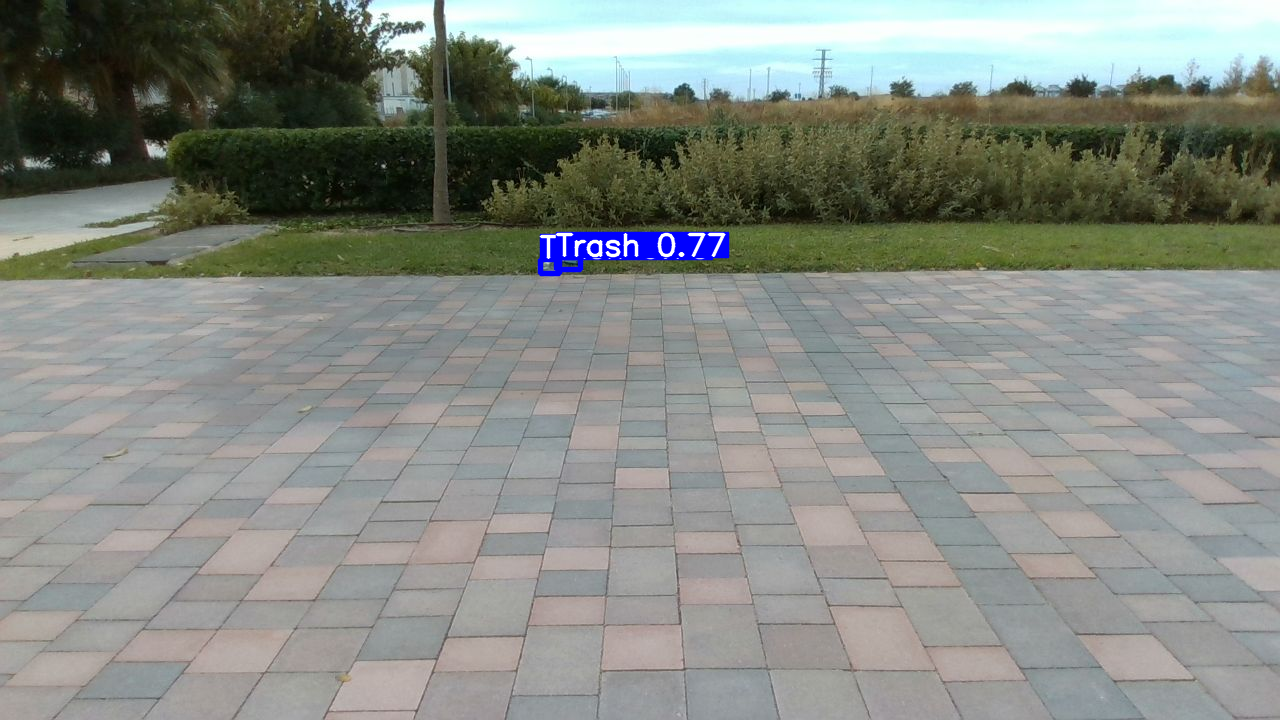}
     \end{subfigure}
    \centering
     \begin{subfigure}
     \centering
         \includegraphics[width=0.23\textwidth]{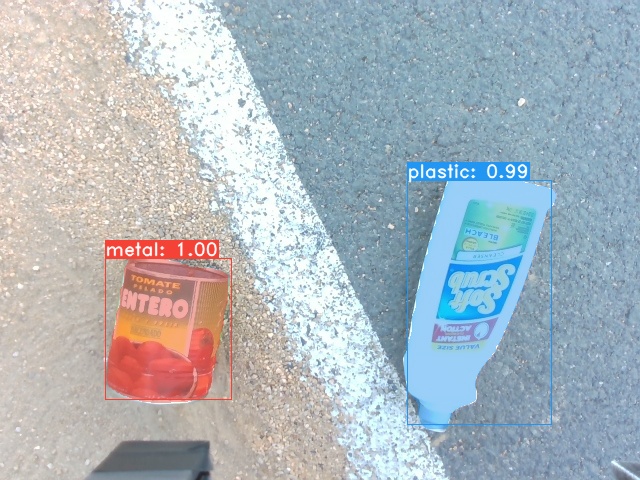}
     \end{subfigure}
     \begin{subfigure}
     \centering
         \includegraphics[width=0.23\textwidth]{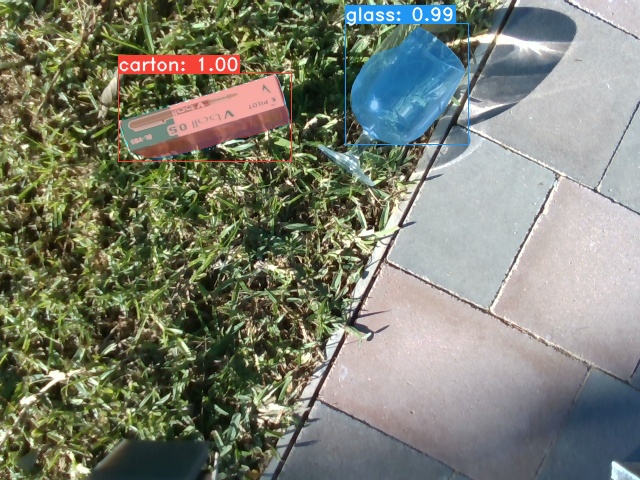}
     \end{subfigure}
    \centering
     \begin{subfigure}
     \centering
         \includegraphics[width=0.235\textwidth]{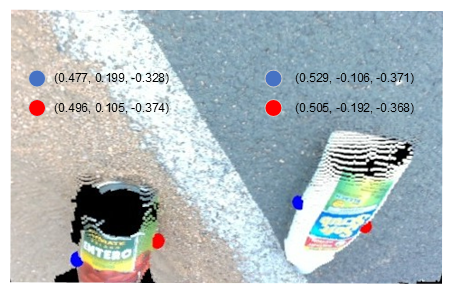}
     \end{subfigure}
     \begin{subfigure}
     \centering
         \includegraphics[width=0.235\textwidth]{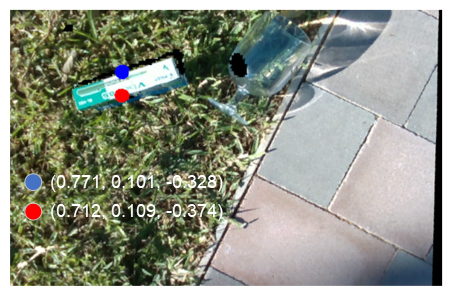}
     \end{subfigure}
     \caption{Localización y navegación (arriba), detección y clasificación (en medio) y estimación de puntos de agarre en residuos (abajo)  \label{fig:navegacion_deteccion_agarre} }
\end{figure}

Después de la evaluación del sistema visual-táctil en modo offline, se ha procedido a evaluarlo en entornos realistas de exteriores y en modo online. Para ello, se realizaron dos experimentos de campo en los que se ejecuta el proceso completo de navegación, detección, reconocimiento y agarre, de 4 objetos y en 2 entornos nunca vistos antes por nuestro sistema visual-táctil. La columna de la izquierda representa el escenario A y la de la derecha el escenario B (Figura \ref{fig:navegacion_deteccion_agarre}). La metodología para cada experimento es la siguiente: se colocan 2 objetos en el suelo simulando que hay residuos. Nuestro robot BLUE, de manera completamente autónoma, detecta los objetos desde la lejanía y navega hacia ellos, realizando una maniobra para situarlos dentro del rango de acción del brazo manipulador. Entonces, se ejecuta el módulo de visión encargado del reconocimiento de los objetos, que los clasifica según el tipo de residuo. Este módulo estima puntos de agarre a partir de la representación de nube de puntos segmentada de los objetos en la escena y determina la pose de agarre para la pinza del manipulador.

\begin{figure}[H]
    \centering
         \includegraphics[width=0.48\textwidth]{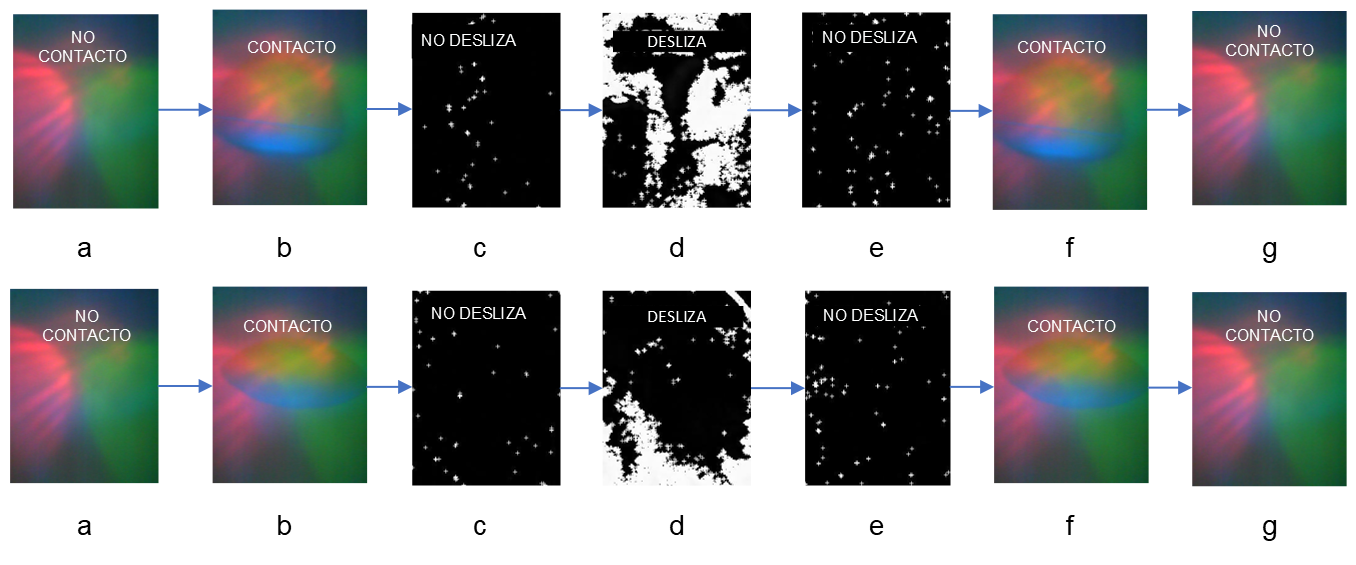}
     
     \caption{Secuencias táctiles durante la tarea de manipulación de residuos. (a, b) corresponden al agarre y detección de contacto. (c, d, e) se obtienen en la detección de deslizamiento y reajuste del cierre de la pinza, y, (f, g) se producen cuando se abre la pinza para depositar el residuo}  \label{fig:secuencia_tactil}
\end{figure}

Después, el manipulador recibe la orden de moverse hasta la posición de agarre y cerrar la pinza hasta que el módulo de percepción táctil, a través del algoritmo de detección de contacto, indica que el objeto ya se encuentra agarrado. Entonces, el brazo manipulador recibe la orden del módulo táctil de levantar y transportar el objeto hacia el deposito de destino (según el tipo de residuo) mientras el algoritmo de deslizamiento corrige la apertura de la pinza en caso de producirse un deslizamiento (Figura \ref{fig:secuencia_tactil}). 

\begin{figure*}
\centering\includegraphics[width=0.9\textwidth, height=10cm]{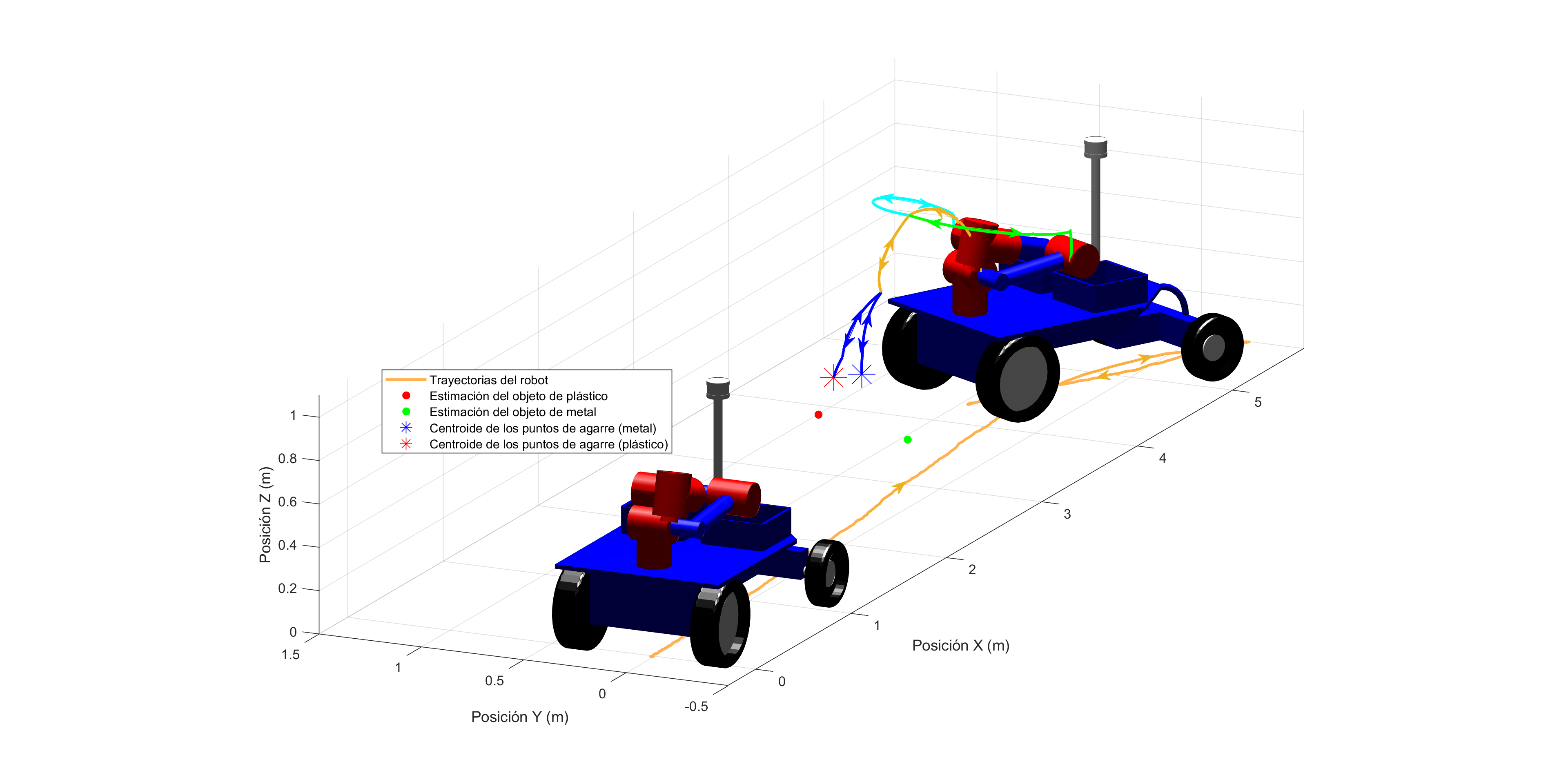}
\caption{Trayectorias de navegación y manipulación generadas para la tarea de recogida de los objetos del escenario A.}  \label{fig:trayectory_3d}
\end{figure*}

\begin{figure*}
\centering
\includegraphics[scale =0.50]{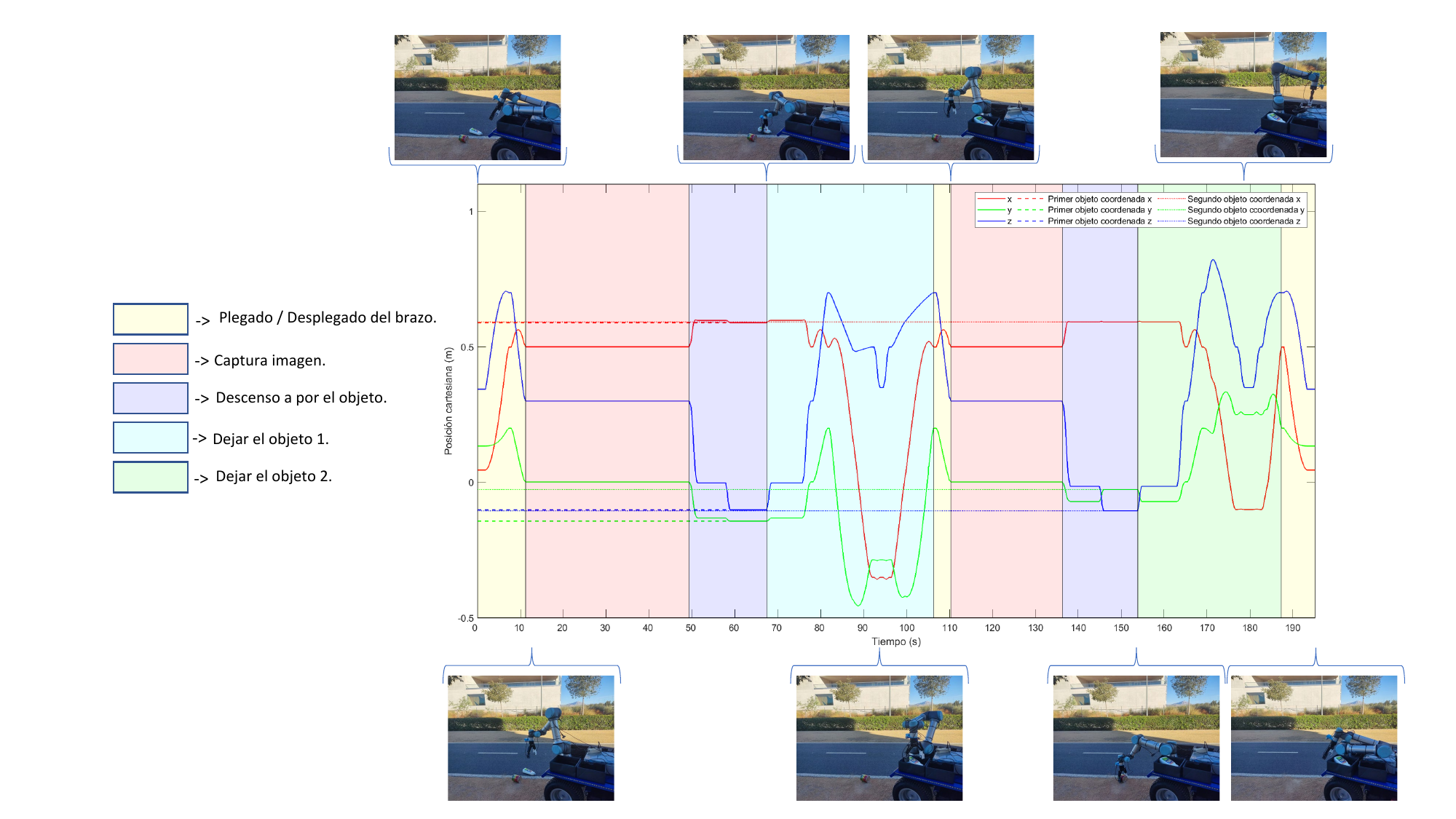}
\caption{Posición de la pinza durante la manipulación de los residuos del escenario A.}  \label{fig:2d_metal_plastic}
\end{figure*}

Finalmente, el brazo manipulador deposita el residuo y vuelve a la posición inicial para volver a ejecutar el proceso de recogida de objetos restantes. Las trayectorias completas de navegación y de manipulación se ilustran en la Figura \ref{fig:trayectory_3d}. Como se observa, el error cometido durante la fase de navegación (puntos rojo y verde) es corregido durante la tarea de manipulación (estrellas roja y azul), consiguiendo así su eliminación. En este ejemplo en particular, se cometió un error de $0.175 \pm 0.06$ m, el cual es un poco menor que el cometido de media en el resto de experimentos realizados únicamente con el módulo de navegación.

La trayectoria cartesiana seguida por el extremo del robot durante la tarea de detección y agarre para el escenario A, se muestra en la Figura \ref{fig:2d_metal_plastic}. En esta figura se muestra, por etapas (zonas de color), el movimiento cartesiano de la pinza durante todo el proceso de posicionamiento para agarre. Las líneas discontinuas indican posiciones deseadas, que vienen dadas por la localización estimada de los objetos por el módulo de visión descrito en secciones previas. Tal y como se observa, el error cometido durante la tarea de navegación llega a anularse, corregido en la fase de estimación de puntos agarre y recogida del objeto.

Todos estos procesos, tanto visuales como táctiles, han sido ejecutados en tiempo real,  con tiempos de cómputo medios en el rango $[30,40]$ ms. en la tarea de localización, $[200, 300]$ ms. en la tarea de reconocimiento, y $[300, 400]$, $[7, 10]$ ms. en detección de contacto y deslizamiento táctil. Esto favorece que todos los métodos y algoritmos puedan ser empleados sin provocar latencia en otros procedimientos embebidos en nuestra plataforma de manipulación móvil.

\section{Conclusiones} 
En este artículo, se ha propuesto un sistema de percepción visual-táctil para la recogida de basura o residuos domésticos no orgánicos, tales como botes, botellas, latas, tetrabricks, tetrapacks, etc.  Nuestro sistema de percepción visual-táctil consta de tres módulos software, dos de ellos con enfoque visual y un tercero con enfoque táctil pero basado en sensores ópticos. Todos los métodos y algoritmos propuestos en el sistema de percepción han sido implementados combinando técnicas de visión por computador y aprendizaje profundo con varias \glspl{cnn}.

La contribución principal se puede resumir de la siguiente manera. Por un lado, se ha diseñado e implementado un método visual para la localización y reconocimiento de residuos sólidos en entorno de exterior. El método combina dos etapas algorítmicas en cascada. En la primera de ellas, se combina una 2D-CNN y datos de geolocalización de un robot móvil para obtener coordenadas geométricas 3D de posición de objetos en la escena. En la segunda, se combina otra 2D-CNN y datos geométricos 3D de la superficie de los objetos para su reconocimiento y posterior estimación del agarre de pinzas de 2 dedos, sin necesidad de una reconstrucción del objeto. El método es flexible y modulable, es decir permite sustituir los modelos de 2D-CNN parametrizados, lo que facilita una arquitectura del sistema de percepción escalable a diferentes robots y entornos. Además, para llevar a acabo estas tareas, se ha creado un dataset específico de residuos domésticos en entornos de exterior. Por otro lado, se ha construido sensores táctiles basados en imagen para controlar el agarre con pinzas robóticas. Este tipo de sensores no dispone de una relación matemática para mapear píxeles a valores de fuerza en N, ni tampoco marcadores que faciliten la estimación de movimiento. Por lo tanto, se han diseñado e implementado algoritmos de detección táctil que nos han permitido desarrollar una estrategia de control para el agarre y reajuste del cierre de la pinza. 

En general nuestro sistema de percepción visual-táctil ofrece buenos resultados, sin embargo también adolece de ciertas limitaciones. Por un lado, el empleo de una única vista RGBD genera en algunos casos, por ejemplo en residuos de vidrio transparente, nubes de puntos 3D sin la necesaria calidad para estimar puntos de agarre adecuados. En estos casos, se puede estudiar el empleo de nubes de puntos más densas obtenidas a partir de la reconstrucción de varias vistas RGBD. Además, se pretende incorporar una cámara multiespectral para analizar como responden los diferentes objetos a diferentes longitudes de onda en aras de favorecer el reconocimiento de nuevos residuos y de mejorar la estimación de puntos de agarre. La cámara multiespectral puede ayudar también a mejorar la precisión en la etapa de localización, ya que es posible que nos permita modificar la detección mediante cuadros delimitadores añadiendo una detección por regiones de segmentación en función de longitudes de onda. 
En relación al módulo táctil para controlar el agarre, se está trabajando en incorporar técnicas de aprendizaje por refuerzo empleando tanto los datos de los sensores táctiles como datos de orientación y apertura de la pinza, para corregir la pose de agarre en caso de detectarse deslizamiento.

\section*{Agradecimientos}

Este trabajo ha sido financiado con Fondos Europeos de Desarrollo Regional (FEDER), el gobierno de la Generalitat Valenciana a través del proyecto PROMETEO/2021/075, y los recursos computaciones fueron financiados a través de la ayuda IDIFEDER/2020/003.

\label{}





\bibliographystyle{elsarticle-harv}
\bibliography{riaibib}







\appendix

\end{multicols}

\end{document}